\documentclass[times, review, 10pt]{elsarticle}

\usepackage[numbers]{natbib}
\usepackage{amsmath,amssymb}
\usepackage{booktabs}
\usepackage{makecell}
\usepackage{multirow}
\usepackage{xcolor}
\usepackage{xspace}
\usepackage{hyperref}
\hypersetup{colorlinks=true}
\def\tsc#1{\csdef{#1}{\textsc{\lowercase{#1}}\xspace}}
\tsc{WGM}
\tsc{QE}

\usepackage{caption}
\usepackage{subcaption}
\usepackage{capt-of}
\usepackage{titlesec}
\usepackage{microtype}
\usepackage{algorithmicx,algorithm}
\usepackage{algpseudocode}
\usepackage[table]{xcolor}

\titlespacing*{\section}{0pt}{1.0ex plus 0.3ex minus 0.2ex}{0.6ex}
\titlespacing*{\subsection}{0pt}{0.8ex plus 0.2ex minus 0.1ex}{0.35ex}
\titlespacing*{\subsubsection}{0pt}{0.5ex plus 0.2ex minus 0.1ex}{0.2ex}

\begin{document}
\definecolor{mypink}{HTML}{FF33A0}
\hypersetup{
    urlcolor=mypink,   
    citecolor=blue,    
    linkcolor=red,     
    pageanchor=false   
}
\setlength{\abovedisplayskip}{4pt plus 1pt minus 1pt}
\setlength{\belowdisplayskip}{4pt plus 1pt minus 1pt}
\setlength{\abovedisplayshortskip}{2pt plus 1pt minus 1pt}
\setlength{\belowdisplayshortskip}{2pt plus 1pt minus 1pt}
\setlength{\textfloatsep}{10pt plus 2pt minus 2pt}
\setlength{\floatsep}{8pt plus 2pt minus 2pt}
\setlength{\intextsep}{8pt plus 2pt minus 2pt}
\renewcommand{\topfraction}{0.90}
\renewcommand{\bottomfraction}{0.80}
\renewcommand{\textfraction}{0.05}
\renewcommand{\floatpagefraction}{0.75}
\let\WriteBookmarks\relax
\def\floatpagepagefraction{1}
\def\textpagefraction{.001}

\newcommand{\tabincell}[2]{\begin{tabular}{@{}#1@{}}#2\end{tabular}}
\definecolor{yyellow}{rgb}{1, 1, 0.7}
\definecolor{oorange}{rgb}{1, 0.85, 0.7}
\definecolor{rred}{rgb}{1, 0.7, 0.7}
\newcommand{\best}{\cellcolor{rred}}
\newcommand{\sbest}{\cellcolor{oorange}}
\newcommand{\tbest}{\cellcolor{yyellow}}

\makeatletter
\patchcmd{\corref}{$\ast\ast$}{$\dagger$}{}{}
\patchcmd{\cortext}{$\ast\ast$}{$\dagger$}{}{}
\makeatother

\begin{frontmatter}

\title{UniFoS: Unifying 3D Foundation Priors with Stereo Geometry for Real-Time Human Gaussian Splatting}

\author[hit]{Shunyuan Zheng}
\ead{sawyer0503@hit.edu.cn}

\author[thu]{Boyao Zhou\corref{lead}}

\author[hit]{Chenyang Wang}

\author[hit]{Kunyi Xie}

\author[thu]{Jiajun Zhang}

\author[pcl]{Boning Liu}

\author[hit]{Xin Sun\corref{cor1}}
\ead{sunxintyc@hit.edu.cn}

\affiliation[hit]{
    organization={Harbin Institute of Technology},
    city={Harbin},
    postcode={150001},
    country={China}
}

\affiliation[thu]{
    organization={Tsinghua University},
    city={Beijing},
    postcode={100084},
    country={China}
}

\affiliation[pcl]{
    organization={Pengcheng Laboratory},
    city={Shenzhen},
    postcode={518055},
    country={China}
}

\cortext[cor1]{Corresponding author.}
\cortext[lead]{Project lead.}

\begin{abstract}
Human novel view synthesis places coupled demands on geometric fidelity, cross-view consistency, and rendering efficiency.
Feed-forward Gaussian Splatting has shown strong potential for this task by avoiding per-scene optimization, yet existing stereo-based methods rely heavily on accurate camera calibration and often degrade in occluded or weakly textured regions.
As a recent breakthrough, 3D foundation models provide strong global geometric priors and support pose-free reconstruction.
However, the predicted depths typically lack fine details and cross-view consistency, while the predicted cameras violate strict epipolar alignment.
In this paper, we propose UniFoS, a feed-forward framework that unifies 3D foundation priors with stereo geometry for real-time human Gaussian Splatting at 2K resolution under both calibrated and uncalibrated settings.
UniFoS first leverages a frozen 3D foundation model to obtain coarse depths and camera predictions, optionally conditioned on input camera poses.
A lightweight iterative depth refinement module then complements the foundation initialization with explicit stereo constraints, converting globally plausible but overly smooth predictions into detail-preserving and view-consistent geometry.
For the uncalibrated setting, we further propose a camera self-correction strategy to improve epipolar consistency without ground-truth camera supervision.
Finally, a lightweight high-resolution Gaussian rendering module predicts pixel-aligned Gaussian maps for photorealistic rendering.
Extensive experiments on synthetic and real-world upper-body human data demonstrate the leading performance of UniFoS on both camera settings.
We develop an immersive display system to further validate the potential of UniFoS in human novel view synthesis applications, achieving 30 FPS and 15 FPS rendering at 2K resolution in calibrated and uncalibrated settings, respectively.
Our project page is available at: \url{https://shunyuanzheng.github.io/UniFoS}.
\end{abstract}


\end{frontmatter}
\hypersetup{pageanchor=true}


\begin{figure}[htpb!]
  \centering
  \vspace{-4mm}
  \includegraphics[width=\textwidth]{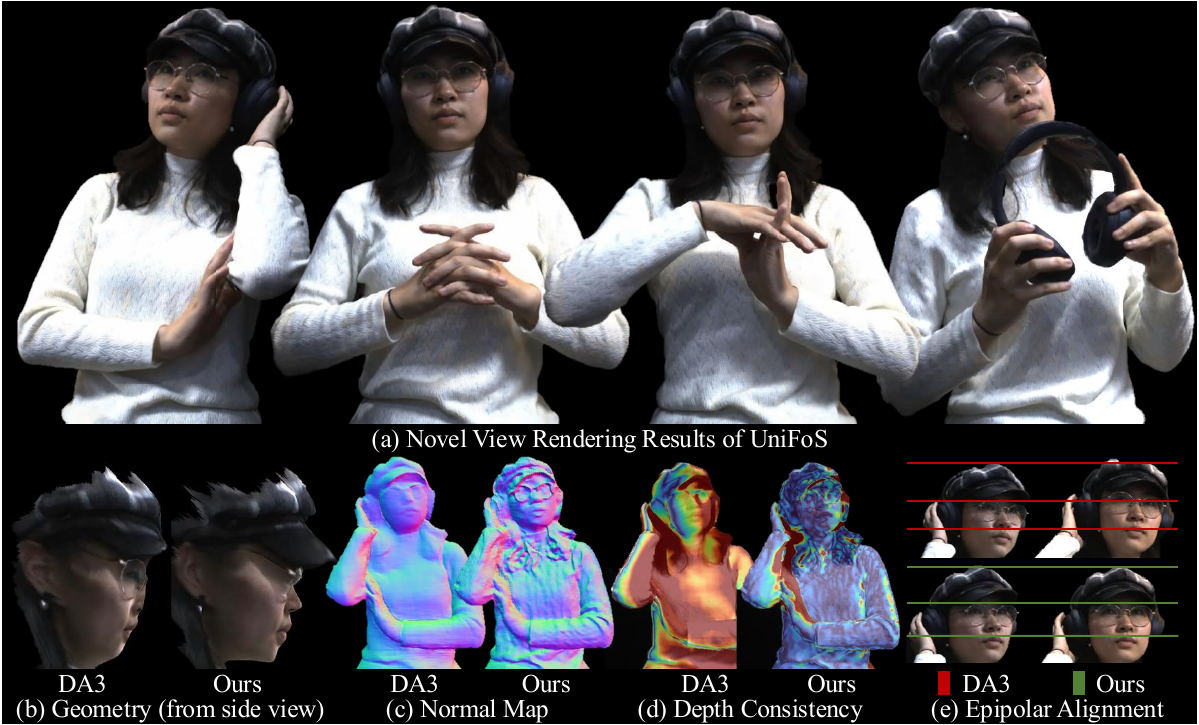}
  \caption{\textbf{Photorealistic real-time rendering and detailed view-consistent geometry.} (a) The proposed UniFoS performs 2K-resolution novel view synthesis at up to 30 FPS. By complementing the predictions of 3D foundation models (\textit{e.g.}, DA3~\cite{lin2026depth}) with explicit geometric constraints, our method recovers (b) fine geometric details (also shown in (c) normal map), improves (d) depth consistency and (e) epipolar alignment. Conversely, 3D foundation models provide reliable initialization and enable UniFoS to work without camera calibration. Note that (d) is evaluated with forward-backward disparity consistency and quantified with pixel error, where a hotter color represents a larger value, and (e) indicates how corresponding pixels deviate from the epipolar lines induced by the predicted cameras.}
  \label{fig:teaser}
\end{figure}

\section{Introduction}
Feed-forward Gaussian Splatting~\cite{charatan2023pixelsplat} has emerged as an efficient and high-quality paradigm for novel view synthesis (NVS), since it avoids the per-scene optimization necessitated by the vanilla 3D Gaussian Splatting (3DGS) methods~\cite{kerbl2023_3dgs}.
In human NVS, a more specific subfield, this technology promotes the innovation of immersive communication~\cite{tu2024tele}, sports broadcasting~\cite{huang2025echoes}, and virtual performance~\cite{sun2025generalizable}.
A typical feed-forward Gaussian Splatting method defines pixel-aligned Gaussian parameters, or termed Gaussian maps, on source-view image planes, which are then unprojected to 3D space using the intermediate depth maps~\cite{charatan2023pixelsplat, zheng2024gpsgaussian} or off-the-shelf depth estimation models~\cite{mescheder2026sharp, xu2025depthsplat}.
Therefore, the accuracy of depth estimation, including its view-consistency under multi-view settings, is critical to the final rendering quality.

Regarding this key issue, early efforts introduce geometric priors by enforcing epipolar constraints~\cite{charatan2023pixelsplat, wewer2024latentsplat} or constructing cost volumes~\cite{zheng2024gpsgaussian, chen2024mvsplat, liu2024mvsgaussian}.
These methods achieve impressive performance when source views are largely overlapped, but degrade significantly in challenging scenarios with occluded or weakly textured regions.
Coarse-to-fine strategies~\cite{liu2024mvsgaussian, zhou2026splat} alleviate this issue to some extent.
Additionally, starting from a coarse geometry narrows the matching candidates and improves inference efficiency.
Recently, 3D foundation models~\cite{wang2024dust3r, leroy2024mast3r, wang2025vggt, lin2026depth, liu2026worldmirror} substantially advanced feed-forward 3D reconstruction by reformulating 3D reconstruction as a pixel-to-scene point map regression task, which enables direct inference of camera parameters, depth, and point tracks simultaneously.
These methods achieve impressive global geometric alignment, especially in the aforementioned challenging cases where traditional multi-view stereo (MVS) struggles.
However, directly applying these geometry-oriented foundation models for rendering tasks yields suboptimal results in detail, especially for high-fidelity human NVS discussed in this paper.
Their predicted depth and cameras are not guaranteed to satisfy the exact pairwise epipolar geometry induced by the observed correspondences.
Errors in camera and point-map predictions compensate for each other, producing globally plausible yet geometrically inconsistent predictions, as shown in Fig.~\ref{fig:teaser}, where the predicted point map lacks details (Fig.~\ref{fig:teaser}~(b)--(c)), exhibits slight cross-view misalignments (Fig.~\ref{fig:teaser}~(d)), and the epipolar geometry induced by the predicted cameras is inconsistent with the cross-view correspondences (Fig.~\ref{fig:teaser}~(e)).

In this paper, we introduce a novel framework, termed UniFoS, to explicitly unify geometric constraints with 3D foundation models via stereo-based lightweight iterative refinement, converting the plausible coarse predictions to renderable geometries with fine-grained details under both calibrated and uncalibrated settings.
We decompose our framework into three components, including coarse geometry reconstruction, lightweight iterative depth refinement, and high-resolution rendering.
This modular design, on one hand, separates reconstruction from rendering, enabling efficient reconstruction that works at low-resolution to support 2K-resolution rendering.
On the other hand, by separating the coarse global geometry reconstruction from local refinement, we can fully exploit the strong priors of pretrained foundation models while maintaining high efficiency through the carefully designed auxiliary refinement module.
The proposed design enables the stereo-based methods to overcome the requirement of depth supervision or even camera calibration, thereby improving generalization to a broader range of real-world scenarios.

Specifically, we first implement an efficient 3D foundation model, which takes source view images as input and camera parameters as an optional condition, to provide coarse geometry prediction including depth map and camera parameters.
When camera poses are conditioned, we align the predictions to the given metric space.
Otherwise, the subsequent modules operate directly in the normalized coordinate space of the foundation predictions.
Herein, with the strong priors of 3D foundation models, we obtain plausible depth maps that are generally view-consistent but suffer from significant smoothness and slight misalignments across views.
To overcome these limitations, inspired by \cite{zheng2024gpsgaussian,lipson2021raft-stereo}, we introduce a lightweight stereo-based iterative geometry refinement module.
Although depth maps predicted by 3D foundation models are insufficient for direct rendering, they serve as an ideal initialization for stereo refinement to greatly narrow the matching candidates, simplifying the depth estimation task with zero initialization to a well-conditioned depth refinement task.
The refinement module compensates for the absence of geometric constraints in foundation models, while foundation models, on the contrary, provide plausible estimates in regions where binocular stereo is degraded.
Finally, the high-resolution rendering module takes source-view images and refined depth maps as input to predict pixel-aligned Gaussian maps through a simplified but effective network for up to 30 FPS real-time photorealistic rendering as shown in Fig.~\ref{fig:teaser}~(a).

It is well known that stereo-based methods rely on accurate camera parameters.
However, under the uncalibrated setting, the only available camera parameters come from the prediction of foundation models, whose errors are inherently coupled with point map prediction.
This issue leads to the following two issues.
First, camera prediction errors cause corresponding pixels to deviate from the epipolar lines induced by the predicted cameras, thereby degrading the subsequent stereo-based refinement.
Second, when directly rendering point maps from source views to novel views, the renders are misaligned with the ground-truth observations, which makes the training dominated by pixel-level losses difficult to converge.
To handle these issues, we introduce a camera self-correction strategy based on bundle adjustment before the refinement module, which does not necessitate the ground-truth camera parameters for supervision.
We correct the predicted camera poses by minimizing the reprojection errors of cross-view feature tracks.
The corrected cameras consequently provide epipolar geometry that is better aligned with the observed correspondences, enabling reliable depth refinement and the unprojection of Gaussian maps in the following.
Therefore, the proposed camera self-correction strategy warrants the effectiveness of the proposed framework in such an uncalibrated situation.

Overall, our contributions are summarized as:
\begin{itemize}
    \item We propose UniFoS, a feed-forward human Gaussian splatting framework that unifies frozen 3D foundation priors with explicit stereo geometry, enabling 2K-resolution real-time human novel view synthesis without depth supervision under both calibrated and uncalibrated settings.
    
    \item We introduce a lightweight foundation-guided iterative depth refinement module, which converts globally plausible but overly smooth foundation depths into detail-preserving and view-consistent geometry for photorealistic Gaussian rendering.
    
    \item We present a camera self-correction strategy for the uncalibrated setting, which requires no ground-truth camera supervision and improves the consistency between the predicted cameras and observed image correspondences, enabling the subsequent depth refinement to maintain reliability.
    
    \item We build an immersive display system to demonstrate the potential of UniFoS in real-time human novel view synthesis applications, which achieves 30 FPS and 15 FPS rendering in calibrated and uncalibrated settings, respectively.

\end{itemize}

\section{Related Work}
\label{sec:related_work}

\subsection{Generalizable 3D Gaussian Splatting}
The explicit 3D Gaussian Splatting (3DGS)~\cite{kerbl2023_3dgs} enables real-time rasterization, and feed-forward variants~\cite{charatan2023pixelsplat, zheng2024gpsgaussian, szymanowicz2024splatter} prevent the time-consuming per-scene optimization with a single network inference.
Splatter Image~\cite{szymanowicz2024splatter} regresses pixel-aligned Gaussians from a single view, and pixelSplat~\cite{charatan2023pixelsplat} exploits epipolar attention to resolve scale ambiguity between two-view inputs.
MVSplat~\cite{chen2024mvsplat} and MVSGaussian~\cite{liu2024mvsgaussian} construct plane-sweep cost volumes for efficient and accurate depth prediction, demonstrating the importance of explicit geometric constraints.
Another line of work, pioneered by Large Reconstruction Models (LRM)~\cite{hong2024lrm}, directly predicts 3D representations with vision transformers~\cite{vaswani2017attention}, and is extended to Gaussian representations~\cite{tang2024lgm}, yet mainly focuses on object-centric datasets due to the capability and computational complexity of the networks.
More recently, DepthSplat~\cite{xu2025depthsplat} and MonoSplat~\cite{liu2025monosplat} leverage priors from pretrained monocular depth estimators~\cite{yang2024depthanything} and design additional adapters to convert relative depth into multi-view-consistent depth.
In this area, the primary issue lies in how to leverage the intermediate depth or underlying geometry proxy, which is critical to the final rendering quality.
Unlike prior work that either builds cost volumes from scratch or adapts monocular priors, UniFoS leverages a 3D foundation model as global geometric initializer and applies explicit stereo refinement to recover renderable, view-consistent geometry.

\subsection{Human-centered Novel View Synthesis}
Human novel view synthesis has long been a central topic due to its wide applications in telepresence~\cite{tu2024tele} and performance capture~\cite{sun2025generalizable}.
Human-centered NVS places particularly stringent requirements on both geometric detail and rendering throughput.
Early implicit-function-based methods~\cite{saito2019pifu} regress pixel-aligned occupancy for clothed human digitization.
NeRF-based approaches~\cite{peng2021neural-body} anchor latent codes on parametric body models~\cite{loper2015smpl} for dynamic humans, but their per-subject optimization and rendering speed hinder interactive applications.
With the advent of 3DGS, per-subject avatar methods~\cite{hu2024gaussianavatar} achieve high-fidelity animatable reconstruction, and relightable extensions~\cite{sun2025generalizable} further support novel illumination.
For generalizable human rendering, GHG~\cite{kwon2024generalizable} regresses Gaussians on the parametric templates~\cite{loper2015smpl}.
EVA-Gaussian~\cite{hu2025eva} adopts a similar insight to LGM~\cite{tang2024lgm} to directly regress multi-view depth and the resulting Gaussian maps with a U-Net-like network under a wide-baseline camera setting.
GPS-Gaussian~\cite{zheng2024gpsgaussian} predicts pixel-wise Gaussian maps via binocular stereo matching, achieving real-time performance.
GPS-Gaussian+~\cite{zhou2025gps} makes an extension to human-scene rendering with rendering-only supervision.
GBC-Splat~\cite{tu2025gbc} combines implicit occupancy estimation with stereo disparity and anchors Gaussians on a reconstructed clothed-human surface.
However, these stereo-based methods rely on accurately calibrated cameras and degrade in occluded or weakly textured regions, and some of them require ground-truth depth for supervision.
In contrast, UniFoS exploits foundation priors to provide plausible initialization in the degraded regions, making it suitable for challenging upper-body human NVS~\cite{tu2024tele} under both calibrated and uncalibrated settings.

\subsection{Feed-forward 3D Foundation Models}
DUSt3R~\cite{wang2024dust3r} reformulates 3D reconstruction as a pixel-to-scene point map regression task, eliminating the requirements of calibrated camera parameters.
MASt3R~\cite{leroy2024mast3r} further grounds image matching in 3D for higher accuracy.
Follow-up work extend this paradigm to multi-view~\cite{yang2025fast3r}, streaming~\cite{ wang2025cut3r}, and dynamic scenes~\cite{zhang2025monst3r}.
VGGT~\cite{wang2025vggt} entirely bypasses optimization steps such as bundle adjustment and simultaneously predicts cameras, depths, point maps, and point tracks in a single forward pass.
The later generation pushes the boundaries of model efficiency~\cite{lin2026depth, wang2026pi} and flexibly supports optional priors as input~\cite{lin2026depth, liu2026worldmirror}.
These reconstruction models inspire a series of NVS methods to obviate the need for camera calibration~\cite{schonberger2016colmap} as pre-processing in conventional pipelines, which is fragile under sparse views with limited overlap.
Splatt3R~\cite{smart2024splatt3r} pioneers pose-free feed-forward NVS by attaching a Gaussian head to MASt3R~\cite{leroy2024mast3r} for rendering.
NoPoSplat~\cite{ye2025nopose} proposes a solution to embed camera intrinsics and predicts all Gaussians in a canonical space directly to resolve scale ambiguity.
AnySplat~\cite{jiang2025anysplat} distills geometry and texture priors from a pretrained VGGT~\cite{wang2025vggt} model within a feed-forward transformer.
DA3~\cite{lin2026depth} and WorldMirror~\cite{liu2026worldmirror}, as unified models, are also equipped with Gaussian heads to support novel view rendering as an application.
In human-centered scenes, HiReFF~\cite{jiang2026hireff} proposes a side-tuning module for 2K resolution rendering while keeping the backbone reconstruction module working on low resolution.
Despite the impressive global geometric alignment in geometry-oriented models, directly using these models for rendering yields suboptimal results.
On one hand, the camera and point map prediction errors are coupled, making the epipolar geometry induced by the predicted cameras inconsistent with the observed cross-view correspondences.
On the other hand, as revealed in \cite{zhao2026structsplat}, their compact 3D features entangle geometry and appearance and are inherently biased toward spatial smoothness, making it difficult to recover fine-grained details by simply attaching an additional rendering head.
Rather than treating foundation models as end-to-end renderers, UniFoS treats their outputs as geometry priors and explicitly completes the missing multi-view constraints required by photorealistic Gaussian rendering.

\section{Method}
\label{sec:method}

\subsection{Problem Formulation and Framework Overview}
\label{sec:method_overview}
\begin{figure}[t]
  \centering
  \includegraphics[width=\textwidth]{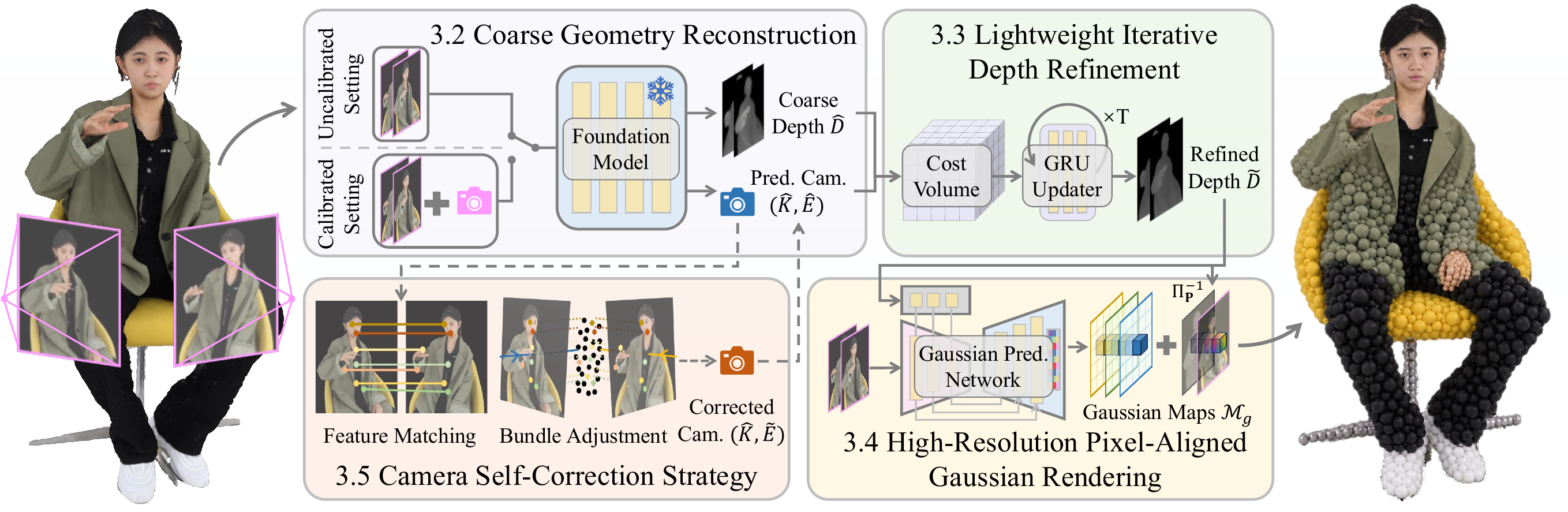}
  \caption{\textbf{Overview of UniFoS.} Our method takes binocular upper-body images as input and their corresponding camera parameters as optional conditions. The input pairs are first fed to a 3D foundation model for coarse depth estimation (Sec.~\ref{sec:foundation_geometry}). Then undergo a lightweight depth refinement module (Sec.~\ref{sec:depth_refinement}) to introduce geometric constraints into foundation predictions. Finally, a high-resolution pixel-aligned rendering module (Sec.~\ref{sec:gaussian_rendering}) regresses pixel-aligned Gaussian maps from the refined depths and source appearance, which are then lifted to 3D space and aggregated for real-time rendering. The camera self-correction (Sec.~\ref{sec:camera_correction}) is necessitated under the uncalibrated setting to ensure the framework remains effective.}
  \label{fig:pipeline}
\end{figure}

The proposed UniFoS, as illustrated in Fig.~\ref{fig:pipeline}, mainly targets high-resolution human novel view synthesis of upper-body performers with real-time feed-forward Gaussian rendering, which has been claimed~\cite{tu2024tele} as a challenging human-centered NVS setup.
Our method takes as input two masked source view images, denoted as $\{\mathbf{I}_{i}\}_{i\in\mathcal{S}}$, where $\mathcal{S}=\{0,1\}$, to reconstruct pixel-aligned Gaussian maps $\mathcal{M}_{g,i}=\{\mathcal{M}_{\boldsymbol{\mu},i},\mathcal{M}_{\mathbf{c},i},\mathcal{M}_{\mathbf{q},i},\mathcal{M}_{\mathbf{s},i},\mathcal{M}_{\alpha,i}\}_{i\in\mathcal{S}}$ defined on source image planes and further unprojected to 3D space as a union Gaussian representation $\mathcal{G}=\{g_i\}_{i\in\mathcal{S}}$. 
These Gaussian maps stand for 3D position map $\mathcal{M}_{\boldsymbol{\mu}}\in\mathbb{R}^{H\times W\times3}$, RGB color map $\mathcal{M}_{\mathbf{c}}\in[0,1]^{H\times W\times3}$, rotation map $\mathcal{M}_{\mathbf{q}}\in\mathbb{R}^{H\times W\times4}$ represents for a unit-quaternion, anisotropic scale map $\mathcal{M}_{\mathbf{s}}\in\mathbb{R}_{+}^{H\times W\times3}$, and opacity map $\mathcal{M}_{\alpha}\in[0,1]^{H\times W\times1}$, where $H \times W$ is the resolution of source images.
We denote the camera $\mathcal{C}_{\mathcal{S}}=\{(\mathbf{K}_{i},\mathbf{E}_{i}) \mid i\in\mathcal{S} \}$ as an optional condition, where $\mathbf{K}_{i}\in\mathbb{R}^{3\times3}$ is intrinsic matrix and $\mathbf{E}_{i}=[\mathbf{R}_{i}\mid\mathbf{t}_{i}] \in\mathbb{R}^{3\times4}$ is extrinsic matrix.

UniFoS consists of the following components.
First, we leverage a 3D foundation model to perform coarse geometry reconstruction (Sec.~\ref{sec:foundation_geometry}), providing a globally coherent but locally smooth depth prediction.
Then, a lightweight iterative depth refinement module (Sec.~\ref{sec:depth_refinement}) introduces explicit constraints into foundation models, recovering fine geometric details and eliminating slight misalignments.
Finally, the high-resolution pixel-aligned rendering module (Sec.~\ref{sec:gaussian_rendering}) structures a carefully designed network to predict pixel-aligned Gaussian maps from the refined depths and source appearance, which are then lifted to 3D space and aggregated for photorealistic real-time rendering.
When input cameras are unavailable, a camera self-correction strategy (Sec.~\ref{sec:camera_correction}) is required before the depth refinement to guarantee that the proposed framework remains effective.
The learnable depth refinement and Gaussian rendering modules are jointly trained through differentiable rendering, without depth supervision (Sec.~\ref{sec:training_objective}).

\subsection{Coarse Geometry Reconstruction}
\label{sec:foundation_geometry}

\noindent\textbf{Coarse Prediction.}
Theoretically, our framework is compatible with a broad class of 3D foundation models.
Since we strive for a general method that can handle both scenarios with and without camera input, and can be applied to immersive display systems, we select Depth Anything 3 (DA3)~\cite{lin2026depth} in practice for its efficiency and flexibility in camera conditions.
The foundation model $\mathcal{F}_{3D}$ takes downsampled source view images as input to predict coarse depth maps $\widehat{\mathbf{D}}^{\downarrow}$ together with camera parameters $\widehat{\mathbf{K}}$ and $\widehat{\mathbf{E}}$:
\begin{equation}
\label{eq:da3_forward}
    \left\{
        \widehat{\mathbf{D}}_{v}^{\downarrow},\widehat{\mathbf{K}}_{v},\widehat{\mathbf{E}}_{v}
    \right\}_{v\in\mathcal{V}}
    =
    \mathcal{F}_{3D}
    \left(
        \{\mathbf{I}_{v}^{\downarrow}\}_{v\in\mathcal{V}};\mathcal{C}_{s}^{\mathrm{in}}
    \right).
\end{equation}
Here, $\mathcal{C}_{s}^{\mathrm{in}}=\{(\mathbf{K}_{i},\mathbf{E}_{i})\}_{i\in\mathcal{S}}$ contains source cameras as condition in the calibrated setting and is empty otherwise.
We set $\mathcal{V}=\mathcal{S}$ when source cameras are given. Otherwise, we additionally include the target views for training supervision in the uncalibrated setting, \textit{i.e.}, $\mathcal{V}=\mathcal{S}\cup\mathcal{T}$, so that their camera predictions can be registered to the same coordinate as the source views.
Only the two source view predictions are used to construct geometry and Gaussians in both camera settings.
The coarse depth maps $\widehat{\mathbf{D}}^{\downarrow}$ are then upsampled to full-resolution $\widehat{\mathbf{D}}$.

\noindent\textbf{Calibrated Metric Alignment.}
Although DA3 accepts camera conditions, its depth prediction remains tied to the internal scale convention of the foundation model.
We therefore estimate a scalar alignment factor $a^\star$ between the predicted and input camera trajectories to rescale the foundation depths.
The DA3~\cite{lin2026depth} leverages the Umeyama algorithm~\cite{umeyama1991least} to predict this scaling factor by treating the camera centers as two point sets.
However, it originally works only for at least three cameras.
In our targeting binocular setup, we augment each two-camera trajectory with a slerp-interpolated middle pose to maintain the effectiveness of this alignment.
Then, coarse source depth in the metric camera frame is $\widehat{\mathbf{D}}/a^\star$, we omit this factor in the following for brevity.
After the depth scale alignment, we replace the predicted $(\widehat{\mathbf{K}},\widehat{\mathbf{E}})$ with the input $(\mathbf{K},\mathbf{E})$.
Note that although the scale is aligned, the conditional input cannot ensure that the predicted rotations are perfectly identical to the given ones.
Even if multi-view depth prediction $\widehat{\mathbf{D}}$ is coherent when unprojected with $(\mathbf{K},\mathbf{E})$, the direct substitution of $(\widehat{\mathbf{K}},\widehat{\mathbf{E}})$ introduces additional errors and slightly degrades the consistency of the multi-view depths.
We leave this remaining inconsistency to the following stereo refinement module.

\subsection{Lightweight Iterative Depth Refinement}
\label{sec:depth_refinement}

\noindent\textbf{Rectification and Foundation Initialization.}
Given the calibrated cameras $(\mathbf{K},\mathbf{E})$, we rectify the two source views into a standard horizontal stereo configuration.
Otherwise, in uncalibrated setting, the cameras $(\widehat{\mathbf{K}},\widetilde{\mathbf{E}})$ predicted by the foundation model and undergone self-correction serve as $(\mathbf{K},\mathbf{E})$.
The self-correction will further be discussed in Sec.~\ref{sec:camera_correction}.
Here, we hypothesize that they conform to the epipolar constraints, which means the correspondences are restricted to horizontal epipolar lines after rectification. 
This yields the rectified images $\mathbf{I}_i^r$, and we reproject the coarse depth estimations to the rectified image plane as $\widehat{\mathbf{D}}^{r}$.
Therefore, all we need to do is to refine a view-consistent disparity (which can be easily converted to depth) between the binocular cameras, thus recovering high-frequency details and correcting the mismatch in foundation depths.

\noindent\textbf{Foundation-guided Iterative Depth Refinement.}
We build our refinement module upon the efficient recurrent stereo design used in GPS-Gaussian~\cite{zheng2024gpsgaussian}, following RAFT-Stereo~\cite{lipson2021raft-stereo}.
The difference is that we transform the zero-initialized prediction to a foundation-guided refinement around a globally plausible initialization, which significantly eases the burden of this module by greatly narrowing the matching candidates.
A shared image encoder first extracts multi-level features from the rectified source views:
\begin{equation}
\label{eq:image_features}
 \{\mathbf{F}_i^s\}_{s=1}^{S}
 =\mathcal{E}_{\mathrm{img}}(\mathbf{I}_i^r),
 \quad
 \mathbf{F}_i^s\in
 \mathbb{R}^{\frac{H}{2^s}\times\frac{W}{2^s}\times d_s},
\end{equation}
where $S=3$ in practice.
Then we use the highest-level feature to build a two-directional correlation volume for iterative lookup.
Instead of estimating the entire correspondence field from scratch, our module treats the foundation-induced disparity as a coarse solution and predicts only its image-supported residual in detail.
At iteration $t$, the current disparity $\mathbf{f}_i^{(t)}$ and its update are written compactly as:
\begin{equation}
\begin{aligned}
\label{eq:residual_recurrent_update}
    \mathbf{f}_i^{(t)}
    &= \mathbf{f}_{i,\mathrm{FM}}^{(0)}+\delta\mathbf{f}_i^{(t)},
    \quad \delta\mathbf{f}_i^{(0)}=\mathbf{0},\\
    \delta\mathbf{f}_i^{(t+1)}
    &= \delta\mathbf{f}_i^{(t)}+\Delta u_i^{(t)},
\end{aligned}
\end{equation}
where the ConvGRU updater predicts $\Delta u_i^{(t)}$ from image context and correlation features sampled around the current correspondence, and $\mathbf{f}_{i,\mathrm{FM}}^{(0)}$ is the disparity converted from foundational depth initialization $\widehat{\mathbf{D}}^{r}$.
The same weights refine both directions in parallel for $T$ iterations, after which learned convex upsampling restores the full-resolution disparity, yielding the depth map $\widetilde{\mathbf{D}}^r$.
The zero-initialized pure regression refinement strategy commonly encounters difficulties in handling large disparity in the first step, and on the other hand, the subsequent steps struggle to generate reasonable displacements towards the correct direction when the first step fails, which is often the case in occluded or weakly textured regions.
The proposed foundation-guided iterative refinement overcomes these drawbacks with the provided initialization, while the learned residual sharpens silhouettes and improves cross-view alignment.
In addition, this stereo-based refinement module is trained without depth supervision via the downstream differentiable Gaussian rendering in Sec.~\ref{sec:gaussian_rendering} and the rendering objective in Sec.~\ref{sec:training_objective}.

\subsection{High-resolution Pixel-aligned Gaussian Rendering}
\label{sec:gaussian_rendering}
On the basis of the fine-grained and view-consistent depth maps, we can now regress pixel-aligned Gaussian maps $\mathcal{M}_g=(\mathcal{M}_{\boldsymbol{\mu}},\mathcal{M}_{\mathbf{c}},\mathcal{M}_{\mathbf{q}},\mathcal{M}_{\mathbf{s}},\mathcal{M}_{\alpha})$.
Following \cite{zheng2024gpsgaussian, hu2025eva, zhou2026splat} the position map $\mathcal{M}_{\boldsymbol{\mu}}$ can be directly determined from depth map via unprojection:
\begin{equation}
\label{eq:gaussian_position}
    \mathcal{M}_{\boldsymbol{\mu},i}(\mathbf{x})
    =
    (\mathbf{R}_i^r)^\mathsf{T}
    \left(
    \widetilde{\mathbf{D}}_i^r(\mathbf{x})
    (\mathbf{K}_i^r)^{-1}\overline{\mathbf{x}}
    -\mathbf{t}_i^r
    \right),
\end{equation}
where $\mathbf{x}=(u, v)^\mathsf{T}$ is the valid pixel in the rectified source image planes, while $\overline{\mathbf{x}}=(u,v,1)^\mathsf{T}$ denotes its homogeneous coordinate.
Moreover, source RGB images serve as the color map $\mathcal{M}_{\mathbf{c}}$.
Therefore, we only need to regress the remaining three Gaussian maps from the image features and the depth features:
\begin{equation}
\begin{aligned}
\label{eq:decoder}
    \mathbf{\Gamma}_i = \mathcal{D}_{parm} (\{\mathbf{F}_i^s\}_{s=1}^{S} \oplus\mathcal{E}_{depth}(\widetilde{\mathbf{D}}_i^r)),\\
    \{\mathcal{M}_{\mathbf{q},i},\mathcal{M}_{\mathbf{s},i},\mathcal{M}_{\alpha,i}\} = \mathcal{D}_{head}(\mathbf{\Gamma}_i \oplus \widetilde{\mathbf{D}}_i^r \oplus {\mathbf{I}}_i^r),
\end{aligned}
\end{equation}
where $\oplus$ is concatenation operation, $\mathcal{E}_{depth}$ stands for depth encoder, $\mathcal{D}_{parm}$ decodes Gaussian features $\mathbf{\Gamma}\in\mathbb{R}^{H \times W \times d_g}$ and further be decoded by $\mathcal{D}_{head}$.
We redesign these regressors to offset the additional cost of introducing 3D foundation models, including a lightweight depth encoder and a shared sub-pixel Gaussian head.

\noindent\textbf{Lightweight Depth Encoder.}
GPS-Gaussian~\cite{zheng2024gpsgaussian} uses a depth encoder that fully mirrors its image encoder.
Since the foundation-guided depth already provides a strong geometric prior, UniFoS designs a pyramid depth encoder with halved channel dimensions and halved convolution layers at each level.
Additionally, we resize the refined depth $\widetilde{\mathbf{D}}^r$ and fuse it with depth features at each level.
This lightweight depth encoder achieves substantially lower inference cost while maintaining comparable reconstruction quality.

\noindent\textbf{Shared Sub-pixel Gaussian Head.}
The former design of the Gaussian head $\mathcal{D}_{head}$ performs full-resolution feature fusion and contains three independent heads $\{h_\mathbf{q}, h_\mathbf{s}, h_\mathbf{\alpha}\}$ for three target Gaussian maps at this resolution.
We find that these repeated heads become the main bottleneck of overhead.
Therefore, we replace them with a shared sub-pixel head whose convolutional decoder terminates at half resolution.
Specifically, we use pixel-unshuffle~\cite{sun2023pixunshuff} to rearrange and downscale the source image and depth as ${\mathcal{P}_{unshuf}(\mathbf{I}}_i^r)\in\mathbb{R}^{H/2 \times W/2 \times 12}$ and $\mathcal{P}_{unshuf}(\widetilde{\mathbf{D}}_i^r)\in\mathbb{R}^{H/2 \times W/2 \times 4}$ in Eq.~\eqref{eq:decoder}, respectively.
We then concatenated them with Gaussian features $\mathbf{\Gamma}^{1/2}\in\mathbb{R}^{H/2 \times W/2 \times d_g}$ at the same resolution.
We implement a shared Gaussian head with a pixel-shuffle~\cite{shi2016pixshuff} operation at the end to regress the combination of three Gaussian maps and rearrange them to the full resolution.
Finally, the combination is separated into independent Gaussian maps with their individual activation function added to satisfy the solution space.

After obtaining all Gaussian maps, we aggregate the valid Gaussians from both source views and use standard differentiable Gaussian splatting~\cite{kerbl2023_3dgs, zheng2024gpsgaussian} to perform high-quality rendering at the desired novel view.
This redesign concentrates representation learning at low and intermediate resolutions while preserving explicit depth boundaries and source texture.

\subsection{Camera Self-correction Strategy for Uncalibrated Settings}
\label{sec:camera_correction}
Camera and depth prediction errors in 3D foundation models may compensate for each other, producing plausible point maps despite inaccurate camera estimates.
However, the epipolar geometry induced by these cameras is not necessarily consistent with the observed cross-view correspondences.
Specifically, corresponding pixels may exhibit non-negligible epipolar residuals or deviate from their predicted epipolar lines after rectification.
This inconsistency violates the assumption made in Sec.~\ref{sec:depth_refinement} and consequently degrades the stereo-based refinement when the predicted cameras $(\widehat{\mathbf{K}},\widehat{\mathbf{E}})$ are used directly in the uncalibrated setting.
Also, using such cameras introduces misalignments between the rendered novel view image and its photometric supervision, making the training difficult to converge.
We address these issues by introducing a camera self-correction strategy, which in advance corrects the predicted camera poses before the depth refinement, preserving the efficiency of the framework.

\begin{algorithm}[t]
\small
\caption{Camera self-correction strategy.}
\label{alg:camera_correction}
\begin{algorithmic}[1]
\Require Images $\{\mathbf{I}_v\}_{v\in\mathcal{V}}$ and foundation cameras
         $\{(\widehat{\mathbf{K}}_v,\widehat{\mathbf{E}}_v)\}_{v\in\mathcal{V}}$; source view indices $\mathcal{S}=\{0,1\}$; optional target view indices $\mathcal{T}=\{2,\ldots,V-1\}$
\Ensure Corrected extrinsics $\{\widetilde{\mathbf{E}}_v\}_{v\in\mathcal{V}}$
\State $\widetilde{\mathbf{E}}_v\gets\widehat{\mathbf{E}}_v$ for all
       $v\in\mathcal{V}$
\State $\mathcal{E}\gets\{(0,v)\}_{v=1}^{V-1}$
       \Comment{star-shaped matching graph}
\State $\mathcal{O}\gets
       \Call{BuildTracks}{\mathcal{E},\{\mathbf{I}_v\},
       \{\widehat{\mathbf{K}}_v,\widehat{\mathbf{E}}_v\}}$
\State $\mathcal{A}\gets\{v \mid N_v(\mathcal{O})\geq n_{\mathrm{cam}}\}_{v=1}^{V-1}$
       \Comment{$n_{\mathrm{cam}}=20$}
\If{$N_{\mathrm{trk}}(\mathcal{O})\geq n_{\min}$ and $\mathcal{A}\neq\varnothing$}
       \Comment{$n_{\min}=30$}
  \State Fix $\widehat{\mathbf{K}}$, $\widehat{\mathbf{E}}_0$, and all poses outside $\mathcal{A}$
  \State $(\{\mathbf{E}_v^\star\}_{v\in\mathcal{A}},b_{src})\gets
         \Call{BundleAdjust}{\mathcal{O},\mathcal{A}}$
  \If{$b_{src}$}
    \State $\widetilde{\mathbf{E}}_v\gets\mathbf{E}_v^\star$ for all
           $v\in\mathcal{A}$
  \EndIf
\EndIf
\State \Return $\{\widetilde{\mathbf{E}}_v\}_{v\in\mathcal{V}}$
\end{algorithmic}
\end{algorithm}

In most foundation models, all predicted cameras are located in the coordinates of the first camera.
We keep this setting to fix the first camera located at the origin with unit rotation, and express all other cameras relative to it as $(\mathbf{R}_v,\mathbf{t}_v)$.
Starting from $\{(\widehat{\mathbf{K}}_v,\widehat{\mathbf{E}}_v)\}_{v\in\mathcal{V}}$, we construct a star-shaped matching graph $\mathcal{E}=\{(0,v)\}_{v=1}^{V-1}$ anchored at the first view.
For each edge, we extract DISK features~\cite{tyszkiewicz2020disk}, match them with LightGlue~\cite{lindenberger2023lightglue}, and reject outliers using RANSAC on the essential matrix.
Matches associated with the same keypoint in the first view are merged into a multi-view track, and we initialize these track points by triangulation with the foundation camera predictions.
At least $n_{min}$ valid tracks are required to invoke the correction.
All cameras supported by at least $n_{cam}$ selected track observations form the active set $\mathcal{A}$ and are optimized jointly, together with the sparse track points $\{\mathbf{X}_m\}$, in a single bundle-adjustment problem~\cite{schonberger2016colmap}:
\par
{
\noindent
\small
\begin{align}
\label{eq:camera_ba}
    \mathcal{L}_{\mathrm{BA}}
    =
    \sum_{(m,v)\in\mathcal{O}}\rho_c\!
    \left(
        \left\|
            \pi\!\left(\widehat{\mathbf{K}}_v,\mathbf{R}_v\mathbf{X}_m+\mathbf{t}_v\right)
            -\mathbf{x}_{mv}
        \right\|_2^2
    \right)
    +
    \sum_{v\in\mathcal{A}}\lambda_v
    \left(
        \frac{\|\mathbf{t}_v\|_2}{\|\widehat{\mathbf{t}}_v\|_2}-1
    \right)^2 ,
\end{align}
}\par
\noindent
where $\mathcal{O}$ contains the retained track--view observations, $\pi$ is perspective projection, $\mathbf{x}_{mv}$ is the observation of track $m$ in view $v$, and $\rho_c$ is the Cauchy robust penalty.
The vector $\widehat{\mathbf{t}}_v$ is the initial relative translation derived from the foundation camera.
Here $\lambda_v=w_v^2/2$, where $w_v$ is the baseline-prior weight assigned according to track support.
We use $w_1=50$ for the source baseline, $w_v=2$ for a well-constrained target, and $w_v=20$ to strengthen the foundation baseline prior when too few tracks are shared across views.
The second term preserves the reliable relative scale of the foundation prediction and removes the scale ambiguity of sparse bundle adjustment~\cite{schonberger2016colmap}.
The intrinsics remain fixed throughout the correction.
Only bounded pose updates that reduce Eq.~\eqref{eq:camera_ba} and retain a sufficient fraction of positive track depths are accepted.
If track support is insufficient or no valid cost-decreasing update is found, the corresponding foundation prediction is retained.
Algorithm~\ref{alg:camera_correction} summarizes the proposed strategy.
This self-correction design requires no ground-truth cameras as supervision.

The correction is invoked only once per inference and optimizes neither network weights nor Gaussians, leaving the learned reconstruction and rendering still feed-forward.
During training, all cameras are jointly corrected with $\mathcal{V}=\mathcal{S}\cup\mathcal{T}$. 
Nevertheless, only the source images and their corrected cameras are used for stereo rectification, depth unprojection, and Gaussian construction, whereas the target views participate solely in camera correction to improve the geometric alignment of the rendering supervision.
At test time, $\mathcal{V}=\mathcal{S}$, and the correction strategy naturally reduces to the single edge $(0,1)$, and the desired novel viewpoints are interpolated between the source views.
Experiments in Sec.~\ref{sec:abla_cam_correction} verify that the resulting discrepancy between the training- and test-time inputs has negligible impact on generalization.
Overall, in the uncalibrated setting, we use $(\mathbf{K},\mathbf{E})=(\widehat{\mathbf{K}},\widetilde{\mathbf{E}})$ in Sec.~\ref{sec:depth_refinement} and Sec.~\ref{sec:gaussian_rendering}.

\subsection{Training Objective}
\label{sec:training_objective}

The depth refinement module and Gaussian regressor are optimized jointly through differentiable rendering~\cite{kerbl2023_3dgs}.
During training, we render the Gaussians to both source and target views as $\widehat{\mathbf{I}}_v$, and compute photometric loss with ground-truth images $\mathbf{I}_v$.
Following \cite{liu2026worldmirror}, a visibility mask $\mathbf{M}^{\mathrm{vis}}_v$ is obtained using frustum projection to mask out the invisible region in novel views.
The masked photometric loss is:
\begin{equation}
\begin{aligned}
\label{eq:rendering_loss}
\mathcal{L}_{mae}
&=
\frac{
\left\|
\mathbf{M}^{\mathrm{vis}}_v \odot \left(\widehat{\mathbf{I}}_v-\mathbf{I}_v\right)
\right\|_1
}{
\left\|\mathbf{M}^{\mathrm{vis}}_v\right\|_1
},\\
\mathcal{L}_{\mathrm{ssim}}
&=
1-\operatorname{SSIM}
\left(
\mathbf{M}^{\mathrm{vis}}_v \odot \widehat{\mathbf{I}}_v,
\mathbf{M}^{\mathrm{vis}}_v \odot \mathbf{I}_v
\right),\\
\mathcal{L}_{\mathrm{render}}
&=
\alpha_1\mathcal{L}_{mae}+\alpha_2\mathcal{L}_{\mathrm{ssim}},
\end{aligned}
\end{equation}
where we set $\alpha_1=0.8$, $\alpha_2=0.2$.
Rendering supervision alone does not uniquely determine the two source geometries.
Analogous to \cite{zhou2025gps}, we regularize the two Gaussian position maps using the symmetric Chamfer distance:
\begin{equation}
\label{eq:chamfer_loss}
\begin{aligned}
 \mathcal{L}_{\mathrm{geo}}
 ={}&
 \frac{1}{|\Omega_0|}
 \sum_{\mathbf{x}\in\Omega_0}
 \min_{\mathbf{y}\in\Omega_1}
 \left\|
 \mathcal{M}_{\boldsymbol{\mu},0}(\mathbf{x})
 -\mathcal{M}_{\boldsymbol{\mu},1}(\mathbf{y})
 \right\|_2^2
 \\
 &+
 \frac{1}{|\Omega_1|}
 \sum_{\mathbf{y}\in\Omega_1}
 \min_{\mathbf{x}\in\Omega_0}
 \left\|
 \mathcal{M}_{\boldsymbol{\mu},1}(\mathbf{y})
 -\mathcal{M}_{\boldsymbol{\mu},0}(\mathbf{x})
 \right\|_2^2 ,
\end{aligned}
\end{equation}
where $\Omega_i$ denotes the valid pixel domain of source view $i$.
We use a stochastic approximation of Eq.~\eqref{eq:chamfer_loss} by sampling $10{,}000$ pixels from each valid domain $\Omega_i$ and retrieving their corresponding 3D positions.
For the uncalibrated setting, we additionally use a perceptual term using LPIPS~\cite{zhang2018lpips} to reduce sensitivity to slight mismatch caused by intrinsic or extrinsic prediction error:
\begin{equation}
\mathcal{L}_{\mathrm{perc}}
=
\operatorname{LPIPS}
\left(
\mathbf{M}^{\mathrm{vis}}_v \odot \widehat{\mathbf{I}}_v,
\mathbf{M}^{\mathrm{vis}}_v \odot \mathbf{I}_v
\right).
\end{equation}
The final training objective is:
\begin{equation}
\mathcal{L} = \gamma_1\mathcal{L}_{\mathrm{render}} + \gamma_2\mathcal{L}_{\mathrm{geo}} + \gamma_3\mathcal{L}_{\mathrm{perc}},
\end{equation}
where we set $\gamma_1=1.0$, $\gamma_2=0.1$ in practice. $\gamma_3$ is required and set to 0.05 only in the uncalibrated setting.

\section{Experiments}
\label{sec:experiments}

\subsection{Datasets and Metrics} 
\label{sec:dataset}
\noindent\textbf{Synthetic data.}
To train and evaluate our method, we render upper-body synthetic data from THuman-sit~\cite{zhang2025ins}, THuman2.0~\cite{yu2021function4d}, and Twindom~\cite{twindom}, which contain 4,722, 526, and 1,779 scans, respectively.
We randomly select 580 scans for validation and the others for training.
For each scan, we position two source view cameras 0.9 meters in front, focusing on the upper body, with intersection angles randomly ranging from 20 to 30 degrees.
Then, randomly place another three cameras on their baselines as novel view cameras.
Each scan is randomly rotated 3 times at the center of the scene for augmentation. We use their SMPL~\cite{loper2015smpl} orientation to ensure face-forward rendering.

\noindent\textbf{Real-world data.}
To evaluate the performance under real-world scenarios, we capture 4 sequences of 1050 frames in total under a similar camera setup.
As shown in Fig.~\ref{fig:live_demo}~(a), we place an additional camera between two source view cameras, highlighted in a green box, to provide novel view images for validation.

\noindent\textbf{Evaluation metrics.}
Following prior work~\cite{zheng2024gpsgaussian, zhou2025gps}, the performance of our method and other baselines is evaluated with PSNR, SSIM~\cite{wang2004ssim}, and LPIPS~\cite{zhang2018lpips} as rendering metrics for the foreground regions in novel views determined by the bounding box of human.
We report end-point-error (EPE) on source-view disparity within the foreground region, measured in pixels, to evaluate geometric accuracy on synthetic data.
We resize the results of all methods to 1,024 resolution for metric evaluation.
For the uncalibrated comparison, since the predicted camera coordinate typically differs from the ground truth in scale and rotation, we could not directly render the novel view images at the desired viewpoints aligned with the ground truth.
Therefore, we follow NoPoSplat~\cite{ye2025nopose} to optimize the target novel view cameras with target view images, adapting the ground-truth camera pose to the predicted camera coordinate.
Note that the predicted source view cameras, reconstructed Gaussians, and network weights remain unchanged during this optimization.
The metrics under this protocol become a target-image-conditioned pose-aligned evaluation.
We conduct the same target pose optimization on all compared methods for fair comparison, and the time cost is not included in the reported runtime.

\subsection{Implementation Details} 
Our method is trained on a single RTX 5880 Ada GPU using AdamW~\cite{adamw} optimizer with an initial learning rate of $1e^{-5}$.
We select the DA3-BASE~\cite{lin2026depth} model as the foundation model in our method for its balance of efficiency and performance.
We keep the foundation model frozen and train the remaining module of the framework for $100k$ iterations with a batch size of 2.
The training takes 18 and 30 hours for calibrated and uncalibrated settings, respectively.

\subsection{Comparisons}
\noindent\textbf{Baselines.}
For calibrated setting, we compare our method with MVS-based methods GPS-GS+~\cite{zhou2025gps} and MVS-Gaussian~\cite{liu2024mvsgaussian}, LRM-based method EVA-GS~\cite{hu2025eva}, monocular depth integrated method DepthSplat~\cite{xu2025depthsplat}, and 3D-foundation-based method WorldMirror~\cite{liu2026worldmirror}.
For uncalibrated setting, we choose MASt3R-based method NoPoSplat~\cite{ye2025nopose}, VGGT-based methods AnySplat~\cite{jiang2025anysplat} and HiReFF~\cite{jiang2026hireff}.
WorldMirror~\cite{liu2026worldmirror} shares a similar network design with VGGT~\cite{wang2025vggt} and enables novel view rendering with both calibrated and uncalibrated inputs.
We train all baseline methods with their open-source code on our upper-body human data on 1-4 RTX 5880 Ada GPUs.

\begin{table*}[t]
\small
\caption{\textbf{Quantitative comparison under calibrated setting.} All methods are evaluated on an RTX 5880 Ada GPU to report the speed of synthesizing one novel view. The \colorbox{rred}{best}, the \colorbox{oorange}{second best}, and the \colorbox{yyellow}{third best} are highlighted with different colors.}
\centering
\scalebox{0.8}{
\begin{tabular}{l|cccc|ccc|c}
\toprule
\multicolumn{1}{c}{\multirow{1}[6]{*}{Method}} & \multicolumn{4}{c}{Synthetic Data} & \multicolumn{3}{c}{Real-world Data} & \multirow{1}[6]{*}{Runtime}\\

\cmidrule{2-8} \multicolumn{1}{c}{} & PSNR$\uparrow$ & SSIM$\uparrow$ & LPIPS$\downarrow$ & \multicolumn{1}{c}{EPE$\downarrow$} & PSNR$\uparrow$ & SSIM$\uparrow$ & \multicolumn{1}{c}{LPIPS$\downarrow$} \\
\midrule

WorldMirror~\cite{liu2026worldmirror}
& 22.01 & 0.802 & 0.208 & 13.28
& 19.48 & 0.762 & 0.207& $131ms@518^2$\\

DepthSplat~\cite{xu2025depthsplat}
& 28.86 & 0.908 & 0.172 & \sbest 2.51
& 23.63 & 0.883 & 0.159 & $95ms@448^2$\\

EVA-GS~\cite{hu2025eva}
& 25.49 & 0.878 & 0.191 & 8.83
& 13.86 & 0.632 & 0.451 & $183ms@1024^2$ \\

MVS-GS~\cite{liu2024mvsgaussian}
& \sbest 29.56 & \tbest 0.928 & \best 0.058 & /
& \tbest 30.34 & \sbest 0.940 & \best 0.038 & $162ms@1024^2$ \\

GPS-GS+~\cite{zhou2025gps}
& \tbest 29.24 & \sbest 0.939 & \tbest 0.104 & \tbest 2.63
& \sbest 30.40 & \tbest 0.935 & \tbest 0.056 & \best $71ms@2048^2$ \\

Ours
& \best 29.63 & \best 0.946 & \sbest 0.072 & \best 1.86
& \best 30.76 & \best 0.956 & \sbest 0.048 & \sbest $76ms@2048^2$ \\

\bottomrule
\end{tabular}}
\label{tab:quant_calib}
\end{table*}

\begin{figure*}[t!]
  \centering
  \includegraphics[width=\textwidth]{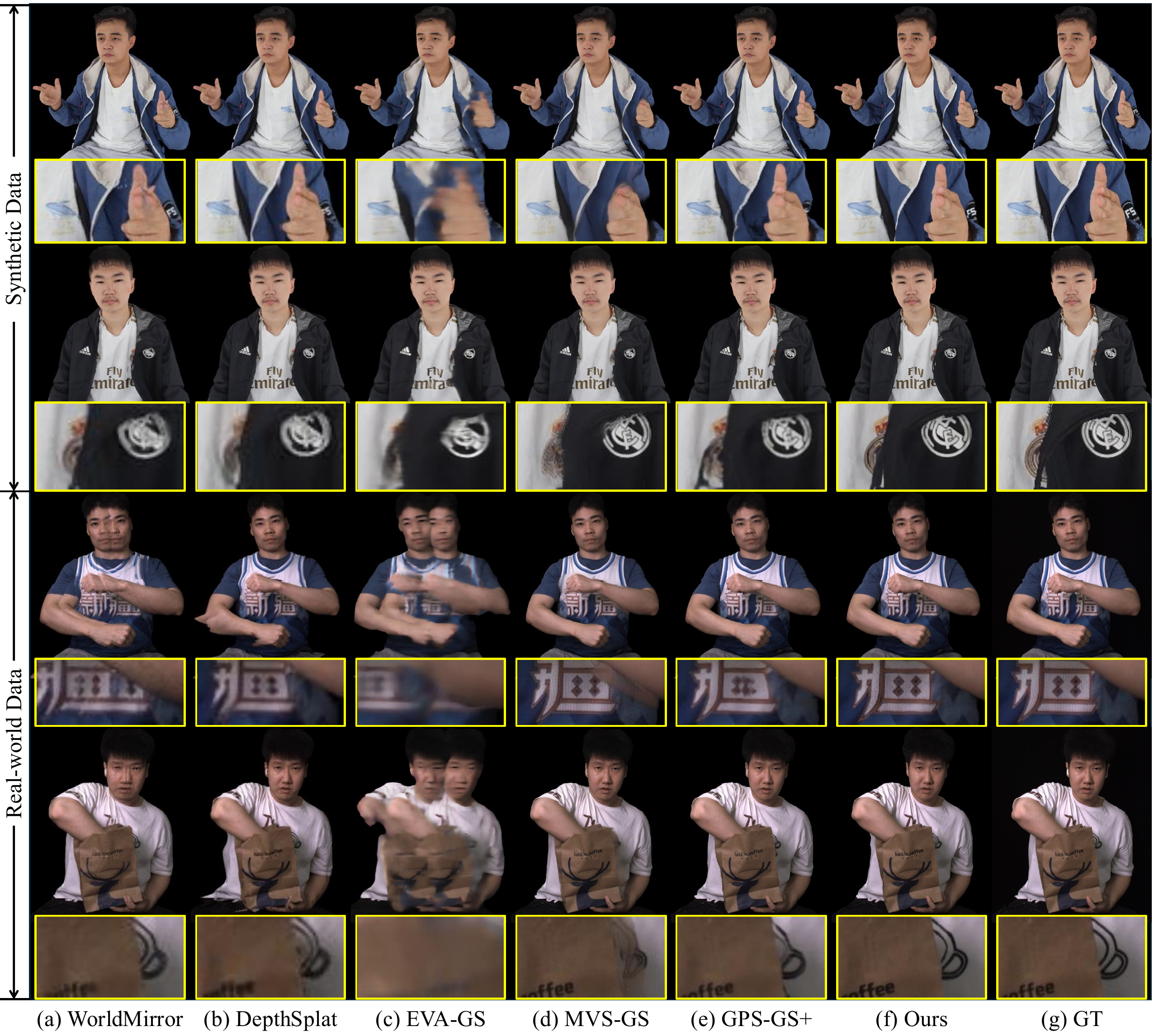}
  \caption{\textbf{Qualitative comparison under calibrated setting.} Our method renders photorealistic human appearances with more accurate geometry.}
  \label{fig:comparison_calib}
\end{figure*}

\subsubsection{Results on Calibrated Setting}

The proposed UniFoS showcases leading geometric accuracy and rendering quality as illustrated in Table~\ref{tab:quant_calib}.
Since MVS-Gaussian~\cite{liu2024mvsgaussian} builds cost volumes and subsequently estimates depth on target views, we could not report EPE results as other methods, which share the same pattern of source view depth estimation and unprojection.
This design bypasses the alignment of Gaussians unprojected from source views in 3D space, producing realistic results in less challenging regions.
However, the performance degrades on occluded regions and thin structures such as fingers, as shown in Fig.~\ref{fig:comparison_calib}~(d).
Regarding the runtime comparison, the lightweight network design offsets most of the additional cost introduced by the 3D foundation model, leading to comparable efficiency compared with GPS-GS+~\cite{zhou2025gps}.
Please refer to Sec.~\ref{sec:abla_network_design} for more analysis.

Qualitative comparison in Fig.~\ref{fig:comparison_calib} shows that our method preserves more details and achieves photorealistic rendering results.
WorldMirror~\cite{liu2026worldmirror} exhibits a larger synthetic-to-real performance drop in our setting, possibly because its camera-conditioned geometry predictor is sensitive to the capture distribution.
Similarly, the monocular depth adapter in DepthSplat~\cite{xu2025depthsplat} and the non-positional encoding design in depth regression of EVA-GS~\cite{hu2025eva} make these two methods overfit to the training camera setup and degrade dramatically in real-world scenes.
Compared with GPS-GS+, the integrated foundation prior in our method brings improvement to depth accuracy, resulting in more faithful rendering of fine structures, especially against self- and object-occlusions.

\subsubsection{Results on Uncalibrated Setting}
\label{sec:exp_nopo}

\begin{table*}[t]
\small
\caption{\textbf{Quantitative comparison under uncalibrated setting.} All methods are evaluated on an RTX 5880 Ada GPU to report the speed of synthesizing one novel view. $\dagger$ NopoSplat~\cite{ye2025nopose} requires additional intrinsic parameters as input and must be manually fixed on Real-world data. The \colorbox{rred}{best}, the \colorbox{oorange}{second best}, and the \colorbox{yyellow}{third best} are highlighted with different colors.}
\centering
\scalebox{0.8}{%
\setlength{\tabcolsep}{3pt}%
\begin{tabular}{l|cccc|cccc|c}
\toprule
\multicolumn{1}{c}{\multirow{1}[6]{*}{Method}} & \multicolumn{4}{c}{Synthetic Data} & \multicolumn{4}{c}{Real-world Data} & \multirow{1}[6]{*}{Runtime}\\

\cmidrule{2-9} \multicolumn{1}{c}{} & PSNR$\uparrow$ & SSIM$\uparrow$ & LPIPS$\downarrow$ & \multicolumn{1}{c}{PSNR$_{src}\uparrow$} & PSNR$\uparrow$ & SSIM$\uparrow$ & LPIPS$\downarrow$ & \multicolumn{1}{c}{PSNR$_{src}\uparrow$}\\
\midrule

WorldMirror~\cite{liu2026worldmirror}
& 20.36 & 0.788 & 0.240 & 21.79
& 21.32 & 0.833 & 0.166 & 23.29 & $129ms@518^2$\\

AnySplat~\cite{jiang2025anysplat}
& 22.64 & 0.817 & 0.247 & \tbest 26.43
& 25.73 & 0.897 & 0.138 & \sbest{29.77} & $116ms@448^2$\\

NoPoSplat~\cite{ye2025nopose}$\dagger$
& \tbest 27.47 & \tbest 0.895 & \tbest 0.176 & \sbest 27.87
& \sbest 30.16 & \sbest 0.945 & \tbest 0.100 & \tbest 29.41 & $125ms@512^2$ \\

HiReFF~\cite{jiang2026hireff}
& \best 27.76 & \sbest 0.904 & \sbest 0.112 & 26.22
& \tbest 29.38 & \tbest 0.941 & \sbest 0.057 & 27.69 & \sbest $197ms@2072^2$ \\

Ours
& \sbest 27.53 & \best 0.912 & \best 0.089 & \best 31.78
& \best 30.80 & \best 0.957 & \best 0.045 & \best 35.85 & \best $147ms@2048^2$ \\

\bottomrule
\end{tabular}}
\label{tab:quant_nopo}
\end{table*}

\begin{figure*}[t]
  \centering
  \includegraphics[width=\textwidth]{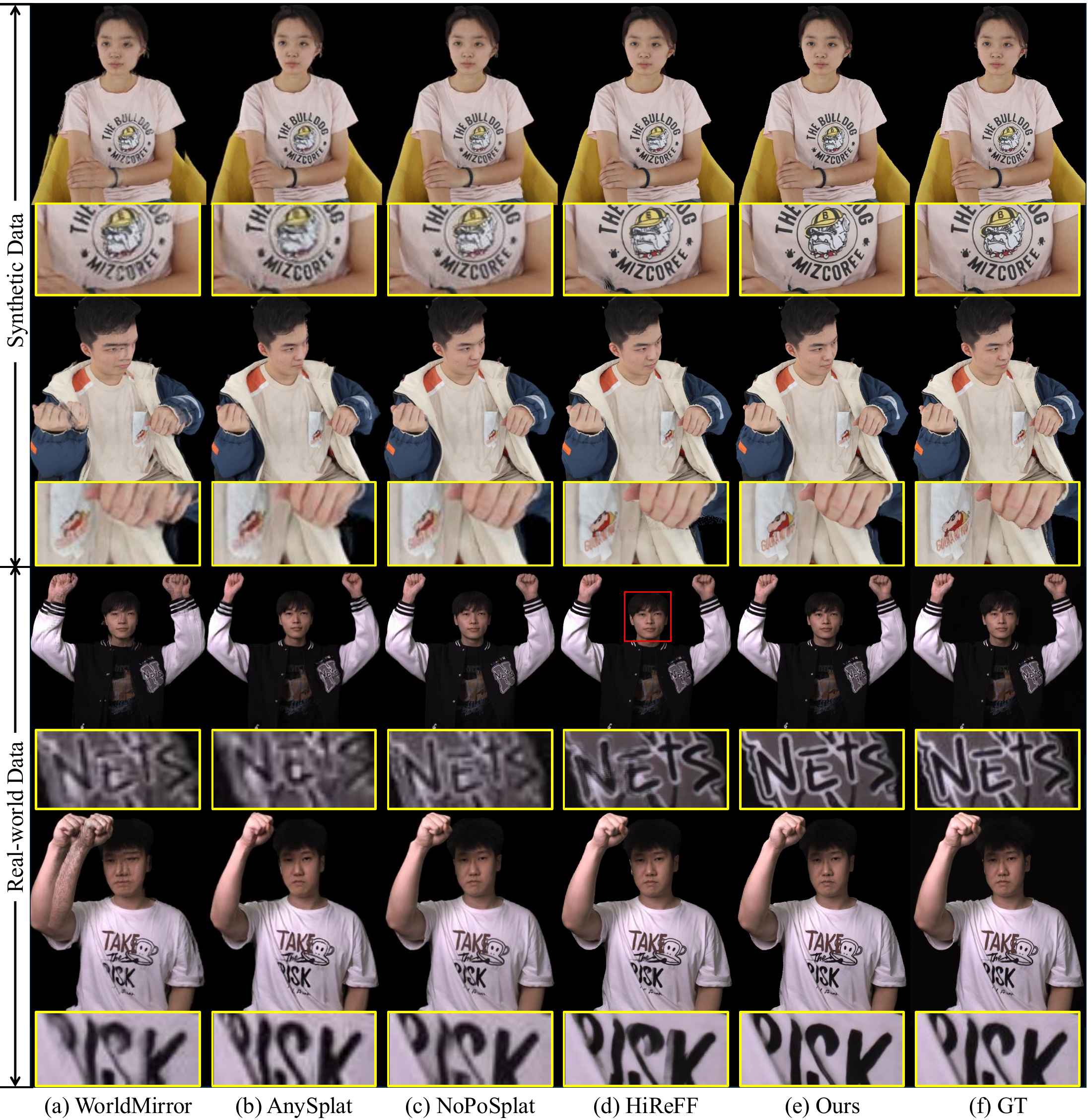}
  \caption{\textbf{Qualitative comparison under uncalibrated setting.} Our method preserves fine-grained details in high resolution with fewer ghosts and noise.}
  \label{fig:comparison_nopo}
\end{figure*}

The quantitative comparison results are shown in Table~\ref{tab:quant_nopo}.
Our method showcases leading rendering quality compared with baseline methods.
Note that the compared methods, except AnySplat~\cite{jiang2025anysplat}, train camera prediction modules with ground truth supervision, while our method leverages a frozen model to provide camera predictions followed by self-correction without using ground truth for supervision.
This design warrants a coherent geometry or Gaussians under the predicted and corrected cameras, leading to high-quality rendering when interpolating novel views.
However, the mismatch between the camera prediction and ground truth typically makes the geometry slightly distorted compared with ground truth.
Although such coherent geometric distortion may have limited perceptual impact for interpolated views, it can disproportionately penalize pixel-aligned metrics such as PSNR and SSIM, making the numerical results less outstanding than the visualization in Fig.~\ref{fig:comparison_nopo}.
We additionally render Gaussians to source views, which do not suffer from geometric distortion, to demonstrate our claim.
We report the PSNR of source views, denoted as PSNR$_{src}$, in Table~\ref{tab:quant_nopo}.
Compared with novel view results, our method and AnySplat achieve significant improvement since we both need no ground-truth camera supervision and suffer from the distortion issue.
On the other hand, some of the other methods degrade on real-world data.
The intrinsic encoding of NoPoSplat~\cite{ye2025nopose} overfits to the training data.
Therefore, we manually change the real focal length to be identical to the synthetic data to produce reasonable results.
Qualitative comparisons are shown in Fig.~\ref{fig:comparison_nopo}.
Our method and HiReFF~\cite{jiang2026hireff} work at high resolution, thus obviously preserving more details than the others.
However, HiReFF suffers from floaters where the depth suddenly changes, resulting in severe effects when the target viewpoint approaches the source views.
Further, its alignment capability decreases in real-world cases, as shown in the boxed face in Fig.~\ref{fig:comparison_nopo}~(d).
Please refer to the supplementary video for more intuitive visualization.

\subsection{Ablation Studies}
In this section, we evaluate the effectiveness of our designed modules. Other than the aforementioned rendering and geometric metrics, we additionally quantify how corresponding pixels deviate from the horizontal epipolar lines using the mean absolute residual vertical disparity (MAVD):
\begin{equation}
E_{MAVD} = \frac{1}{N}\sum_{i=1}^{N}\left|y_i^{L}-y_i^{R}\right|,
\label{eq:mavd}
\end{equation}
where $(y_i^{L}, y_i^{R})$ are the vertical coordinates of the $i$-th corresponding pixel pairs in rectified stereo images.
We randomly select $N=10,000$ unoccluded pixel pairs per image according to the ground-truth disparities.
All ablations are trained and tested on the proposed synthetic data.

\subsubsection{Effects of Depth Refinement Module}
\begin{figure*}[t]
  \centering
  \vspace{2mm}
  \includegraphics[width=\textwidth]{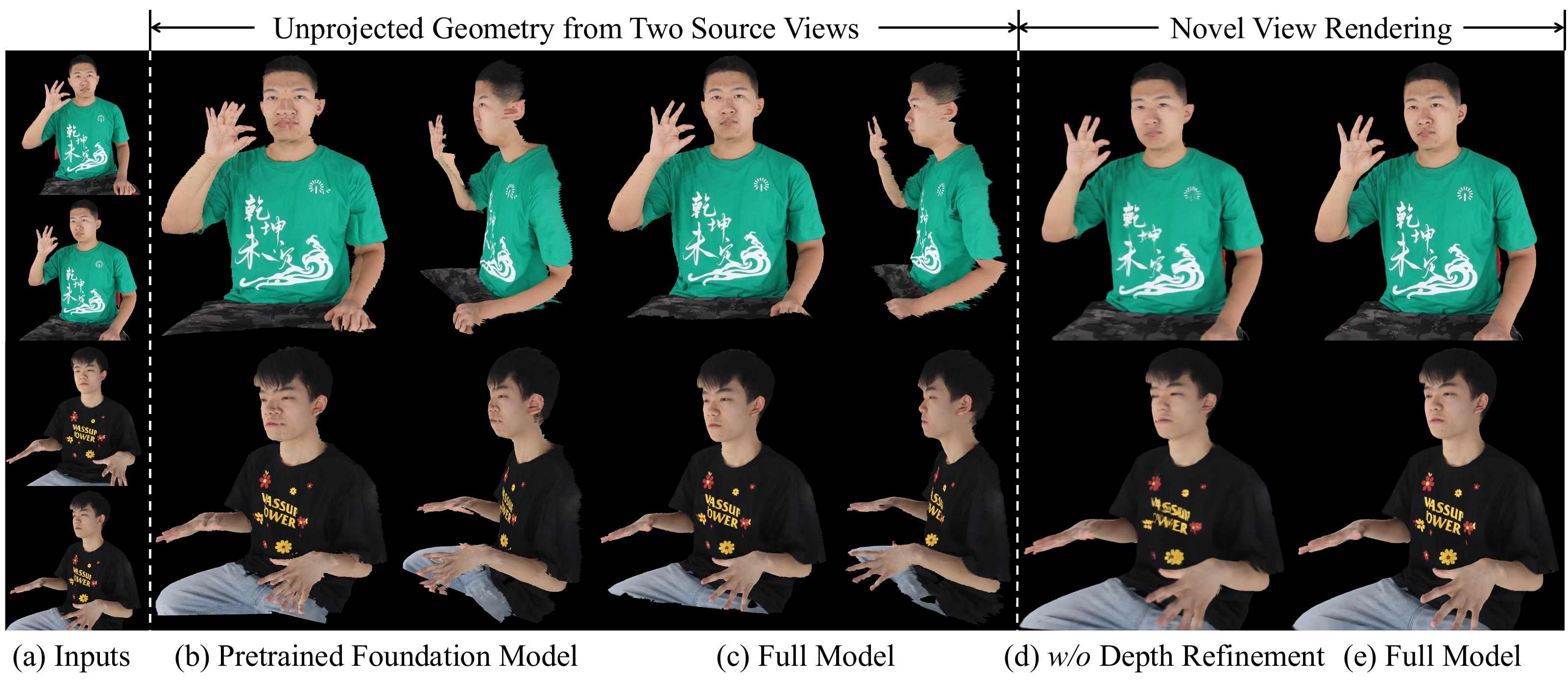}
  \caption{\textbf{Qualitative ablation study on the designed depth refinement module.} The geometries are shown with colored mesh, displayed from front and side views. The proposed design enables the unprojected source view depth to be more coherent, leading to more visually appealing novel view rendering results.}
  \vspace{2mm}
  \label{fig:abla_depth_refine}
\end{figure*}

We design a variant without the depth refinement module and add a depth residual head alongside other Gaussian maps, which provides trainable depth residuals added to the pretrained foundation depths for rendering.
The geometric comparison results in Fig.~\ref{fig:abla_depth_refine} (b) and (c) show that the depth refinement module enhances multi-view depth consistency, which can also be reflected in EPE values manifested in Table~\ref{tab:ablation_depth_refine}.
As shown in Fig.~\ref{fig:abla_depth_refine} (d) and (e), directly using the pretrained foundation depth causes blurry renderings, while our refined depth avoids this issue, improving the rendering quality significantly, which can also be revealed in Table~\ref{tab:ablation_depth_refine}.

\begin{table*}[t]
\small
\caption{\textbf{Quantitative ablation study on the designed depth refinement module.} We use DA3-BASE~\cite{lin2026depth} as the pretrained foundation model.}
\centering
\scalebox{0.9}{
\begin{tabular}{l|ccccc}
\toprule
Method & PSNR$\uparrow$ & SSIM$\uparrow$ & LPIPS$\downarrow$ & EPE$\downarrow$ \\
\midrule

Pretrained Foundation Model
& / & / & / & 17.32 \\

\textit{$w/o$} Depth Refinement
& 24.70 & 0.866 & 0.174 & 8.23 \\

Full Model
& \textbf{29.63} & \textbf{0.946} & \textbf{0.072} & \textbf{1.86}\\

\bottomrule
\end{tabular}}
\label{tab:ablation_depth_refine}
\end{table*}

\vspace{2mm}
\subsubsection{Effects of Camera Self-correction Strategy}
\label{sec:abla_cam_correction}
\begin{figure*}[t]
  \centering
  \includegraphics[width=\textwidth]{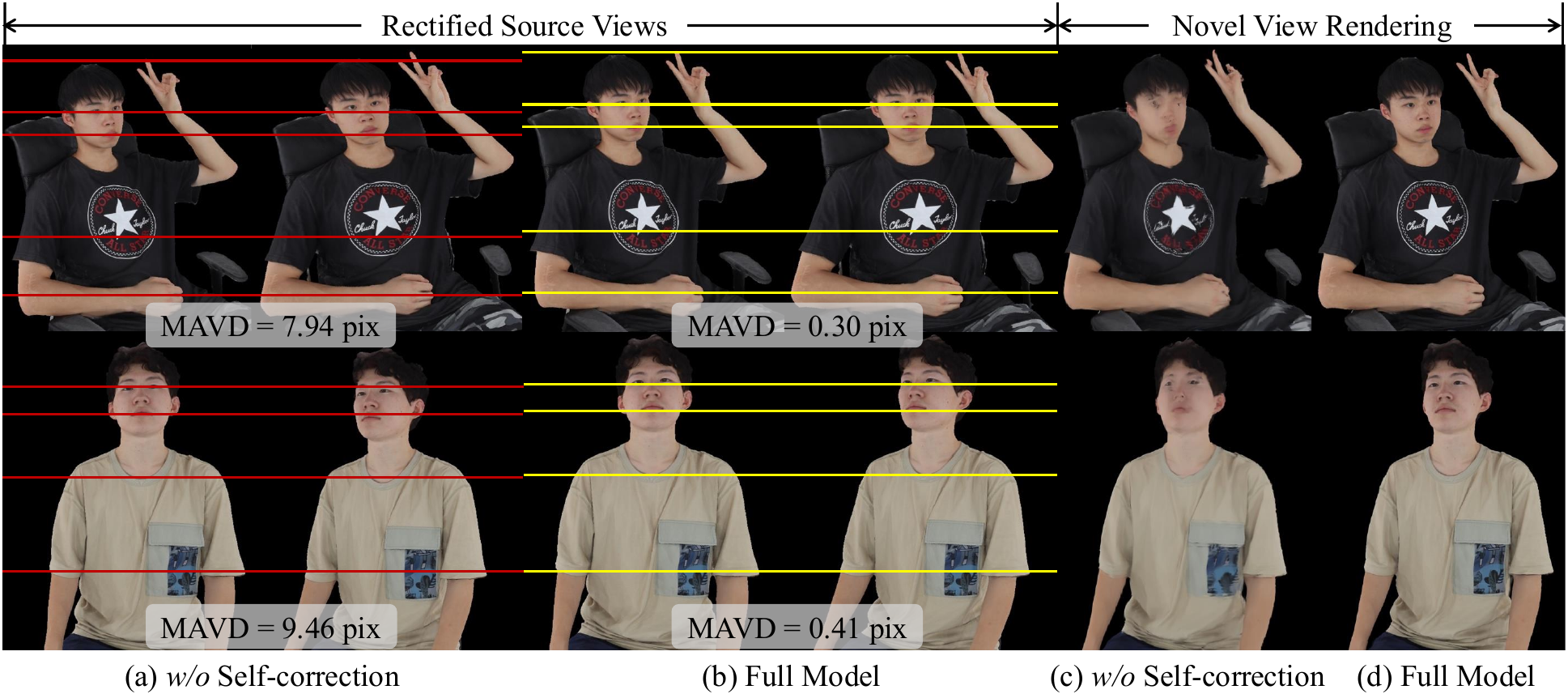}
  \caption{\textbf{Qualitative ablation study on the designed camera self-correction strategy.} MAVD quantifies how corresponding pixels deviate from the horizontal epipolar lines on the rectified source images. The proposed strategy ensures that the epipolar geometry is well aligned with the correspondences, avoiding ghosting artifacts when synthesizing novel views.}
  \label{fig:abla_cam_correction}
\end{figure*}

First, we compare our full model with an ablated model without camera self-correction, which directly leverages the predicted cameras from the foundation model to rectify the image pairs and unproject Gaussians.
The corresponding pixel pairs in Fig.~\ref{fig:abla_cam_correction}~(a) deviate from the epipolar lines induced by the predicted cameras, which can be observed near eyes, mouth, fingers, and the patterns on the clothing.
Synthesizing novel views with these cameras causes severe ghosting artifacts on the mentioned body regions, as shown in Fig.~\ref{fig:abla_cam_correction}~(c).
The proposed camera self-correction strategy addresses these issues with decreased MAVD and improved rendering quality, which is further verified in Table~\ref{tab:ablation_cam_correction}.

\begin{table*}[t]
\small
\caption{\textbf{Quantitative ablation study on the designed camera self-correction strategy.} MAVD refers to mean absolute vertical disparity, measured in pixels. The 2-view correction is our default inference setting adopted in all other experiments.}
\centering
\scalebox{0.9}{
\begin{tabular}{l|ccccc}
\toprule
Method & PSNR$\uparrow$ & SSIM$\uparrow$ & LPIPS$\downarrow$ & MAVD $\downarrow$\\
\midrule

\textit{$w/o$} Self-correction
& 25.18 & 0.861 & 0.139 & 5.159\\

Full Model (5-view correction)
& \textbf{27.59} & 0.910 & \textbf{0.089} & 1.521\\

Full Model (2-view correction, default)
& 27.53 & \textbf{0.912} & \textbf{0.089} & \textbf{1.510}\\

\bottomrule
\end{tabular}}
\label{tab:ablation_cam_correction}
\end{table*}

Then, we conduct another ablation on the number of input views when implementing camera self-correction.
In the training phase, we feed both source and novel view images to the foundation model and perform self-correction on the reasoning camera predictions.
However, at test time or in real-world scenes, we can only take source images as input to interpolate novel views, which may cause potential bias in the self-correction strategy.
Therefore, we additionally evaluate our full model with all views input, \textit{i.e.}, 5 views in our dataset, which becomes identical to the training phase.
The results in Table~\ref{tab:ablation_cam_correction} reveal that involving novel view images for inference brings negligible additional improvement.
In other words, the discrepancy between the training- and test-time inputs has nearly no effect on the camera self-correction strategy.

\subsubsection{Effects of Lightweight Rendering Network Design}
\label{sec:abla_network_design}

\begin{table*}[t]
\small
\caption{\textbf{Quantitative ablation study on the designed lightweight rendering network.} ``\textit{Light.~Dep.~Enc.}'' represents lightweight depth encoder, and ``\textit{SSP~GS~Head}'' represents shared sub-pixel Gaussian head.}
\centering
\scalebox{0.85}{
\begin{tabular}{c|cc|ccccc}
\toprule
Variant & \textit{Light.~Dep.~Enc.} & \textit{SSP~GS~Head} & PSNR$\uparrow$ & SSIM$\uparrow$ & LPIPS$\downarrow$ & EPE$\downarrow$ & Runtime$\downarrow$ \\
\midrule

(a) & &
& 29.43 & 0.944 & 0.074 & 1.92 & 91~$ms$\\

(b) & $\checkmark$ &
& \textbf{29.64} & \textbf{0.947} & \textbf{0.070} & 1.87 & 85~$ms$ \\

(c) & & $\checkmark$
& 29.61 & 0.946 & 0.071 & 1.89 & 82~$ms$ \\

Full Model& $\checkmark$ & $\checkmark$
& 29.63 & 0.946 & 0.072 & \textbf{1.86} & \textbf{76~$ms$} \\

\bottomrule
\end{tabular}}
\label{tab:ablation_network_design}
\end{table*}

We carefully redesign two components of the rendering network, namely the lightweight depth encoder and the shared sub-pixel Gaussian head, to offset the latency (20~$ms$ for DA3-BASE) of integrating the foundation model, preserving high efficiency without sacrificing performance.
Table~\ref{tab:ablation_network_design} quantifies the contribution of each lightweight design.
Compared with the baseline variant (a), introducing the lightweight depth encoder (b) and shared sub-pixel Gaussian head (c) reduces the runtime by 6~$ms$ and 9~$ms$, respectively.
Combining both designs, the full model saves 15~$ms$ compared with the baseline, thereby offsetting most of the 20~$ms$ overhead introduced by the foundation model.
Meanwhile, it retains comparable performance, demonstrating that the proposed lightweight designs effectively balance quality and efficiency.

\subsubsection{Sensitivity to 3D Foundation Models}

The proposed method is compatible with a broad class of 3D foundation models.
In Table~\ref{tab:ablation_foundation_model}, we report the quantitative results and the corresponding runtime when substituting various foundation models in our pipeline.
Our method is robust to different integrated models, while WorldMirror~\cite{liu2026worldmirror} showcases its superiority in both settings.
However, we select DA3-BASE as our default model for its balance between performance and efficiency.
From another perspective, in the upper-body human NVS task, integrating WorldMirror into our rendering framework achieves superior quality compared to directly attaching a Gaussian head for rendering, which is its original design for rendering tasks, as reported in Table~\ref{tab:quant_calib} and Table~\ref{tab:quant_nopo}.

\begin{table*}[t]
\small
\caption{\textbf{Quantitative ablation study on the sensitivity to 3D foundation models.} VGGT~\cite{wang2025vggt} only supports uncalibrated inference. *DA3-BASE is the default model we used.}
\centering
\scalebox{0.8}{
\begin{tabular}{l|ccccc|cccc}

\toprule
\multicolumn{1}{c}{\multirow{2}{*}{\makecell{Foundation\\Model}}} & \multicolumn{5}{c}{Calibrated Setting} & \multicolumn{4}{c}{Uncalibrated Setting}\\

\cmidrule{2-10} \multicolumn{1}{c}{} & PSNR$\uparrow$ & SSIM$\uparrow$ & LPIPS$\downarrow$ & EPE$\downarrow$ & \multicolumn{1}{c}{Runtime$\downarrow$} & PSNR$\uparrow$ & SSIM$\uparrow$ & \multicolumn{1}{c}{LPIPS$\downarrow$} & Runtime$\downarrow$\\
\midrule

DA3-SMALL
& 29.49 & 0.942 & 0.075 & 1.99 & 69~$ms$
& 27.21 & 0.906 & 0.110 & 142~$ms$ \\

DA3-BASE*
& 29.63 & 0.946 & 0.072 & 1.86 & 76~$ms$
& 27.53 & 0.912 & 0.089 & 147~$ms$\\

DA3-LARGE
& 29.75 & 0.948 & 0.071 & 1.80 & 118~$ms$
& 27.47 & 0.911 & 0.083 & 193~$ms$ \\

WorldMirror
& 29.84 & 0.949 & 0.070 & 1.69 & 151~$ms$
& 27.74 & 0.915 & 0.081 & 214~$ms$ \\

VGGT
& / & / & / & / & /  
& 26.94 & 0.901 & 0.086 & 201~$ms$ \\

\bottomrule
\end{tabular}}
\label{tab:ablation_foundation_model}
\end{table*}

\section{Live Demo System}
\label{sec:live_demo}

\begin{figure*}[t]
  \centering
  \includegraphics[width=\textwidth]{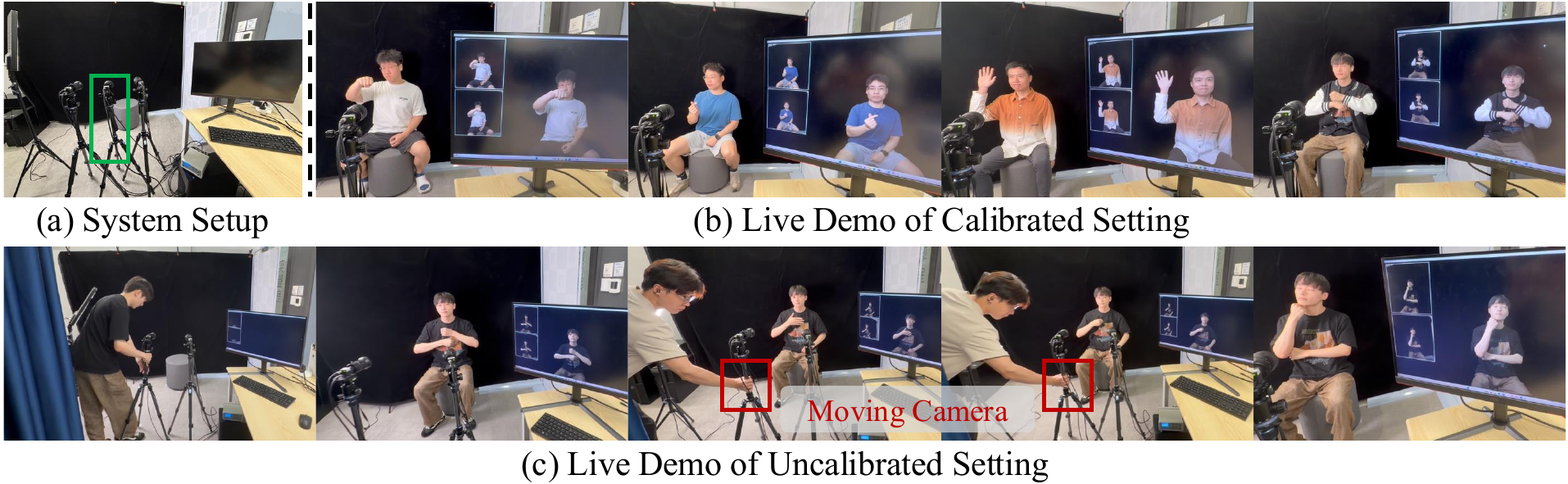}
  \caption{\textbf{Camera System Setup and Illustration of Live Demo.} We use the camera rigs in (a) for collecting real-world test data and capturing image streams for the live demo. The camera highlighted in green box serves as a novel viewpoint, only used in preparing offline data. Live Demo results under calibrated and uncalibrated settings are illustrated in (b) and (c), respectively. The cameras in (c) are casually placed, and one of them keeps moving.}
  \label{fig:live_demo}
\end{figure*}

We build an immersive display system to demonstrate the potential of UniFoS in high-quality real-time human NVS applications.
The live demo results of both camera settings are illustrated in Fig.~\ref{fig:live_demo} (b) and (c).
Our system performs high-quality upper-body novel view synthesis of any incoming performer with challenging motions.
We keep moving the input camera as highlighted in red box in Fig.~\ref{fig:live_demo}~(c) to show the robustness under uncalibrated setting.
The synthesized novel views maintain photorealism against the moving input views.
Please refer to the supplementary video for comprehensive live demo presentations.

\begin{table*}[t]
\small
\caption{\textbf{Runtime Analysis.} The PyTorch version is evaluated on RTX 5880 Ada GPU and TensorRT on RTX 5090D v2 GPU. ``Depth Ref.'' stands for Depth Refinement module and ``GS Reg.'' for Gaussian Regression module. Self-Correction strategy is only used in uncalibrated setting. Therefore, we report the calibrated/uncalibrated overall runtime, respectively.}
\centering
\scalebox{0.9}{
\begin{tabular}{l|cc|cc}
\toprule
\multicolumn{1}{c}{\multirow{1}[6]{*}{Module}} & \multicolumn{2}{c}{PyTorch} & \multicolumn{2}{c}{TensorRT}\\

\cmidrule{2-5} \multicolumn{1}{c}{} & Runtime & \multicolumn{1}{c}{Precision} & Runtime & \multicolumn{1}{c}{Precision} \\
\midrule

Human Matting
& 14~$ms$ & FP32 & 7~$ms$ & FP32 \\

Foundation Model
& 20~$ms$ & BF16+FP32 & 5~$ms$ & FP16 \\

Self-Correction (in Uncalib.)
& 71~$ms$ & FP32 & 30~$ms$ & FP16+FP32 \\

Depth Ref. + GS Reg.
& 56~$ms$ & FP32 & 17~$ms$ & FP32 \\

Overall
& 90~$ms$/161~$ms$ & / & 29~$ms$/59~$ms$ & / \\

\bottomrule
\end{tabular}}
\label{tab:runtime}
\end{table*}

We use an RTX 5090D v2 graphics card to implement the live demo system.
Our system is implemented with TensorRT in C++ for fast inference.
The runtime analysis is listed in Table~\ref{tab:runtime}.
Note that the foundation model, namely DA3-BASE, autocasts the backbone to BF16 and keeps the camera and depth head in FP32 in PyTorch.
In the camera self-correction module, we use FP16 for feature extraction and matching, while using FP32 for bundle-adjustment-related operations.
The calibrated system achieves 30 FPS real-time efficiency at 2K resolution, while the uncalibrated version operates at 15 FPS, and both interactively and seamlessly perform high-quality human NVS.

\section{Discussion}
\label{sec:conclusion}

\noindent\textbf{Conclusion.}
In this paper, we present UniFoS, a feed-forward framework that unifies 3D foundation priors with stereo geometry for real-time high-resolution human Gaussian Splatting.
UniFoS uses a frozen 3D foundation model to provide globally plausible depth and camera predictions, and complements these predictions with explicit geometric constraints through a lightweight iterative depth refinement module, which performs foundation-guided stereo refinement, attaching local geometric details and improving cross-view consistency.
To support the uncalibrated setting, we further introduce a camera self-correction strategy that corrects the predicted camera poses without requiring ground-truth camera supervision, thereby making the induced epipolar geometry better aligned with observed image correspondences.
With the refined geometry, a lightweight rendering network predicts pixel-aligned Gaussian maps for photorealistic 2K novel view synthesis.
Extensive experiments on synthetic and real-world upper-body human data demonstrate that UniFoS achieves high-quality rendering and accurate geometry under both calibration settings.
Finally, we build an immersive display system to validate the practical applicability of the proposed UniFoS, achieving 30 FPS and 15 FPS rendering at 2K resolution in calibrated and uncalibrated settings, respectively.

\noindent\textbf{Limitations.}
Although our uncalibrated version outperforms the baseline methods on high-resolution rendering efficiency, its latency increases significantly compared with our calibrated version, preventing fully real-time inference in the uncalibrated setting on a single graphics card.
The main reason lies in the redundancy in feature extraction when implementing camera self-correction.
Further, as discussed in Sec.~\ref{sec:exp_nopo}, the geometric distortion issue compromises pixel-aligned quantitative evaluation, making our numerical improvements appear less pronounced than the clear gains observed in qualitative comparisons.
Future work may explore lightweight feature sharing and carefully regularized human-centered adaptation of 3D foundation models to further improve efficiency and geometric consistency.











\bibliographystyle{elsarticle-num}

\bibliography{cas-refs}

@String(PAMI = {IEEE Trans. Pattern Anal. Mach. Intell.})

@String(CVPR= {IEEE Conf. Comput. Vis. Pattern Recog.})

@String(ICCV= {Int. Conf. Comput. Vis.})

@String(ECCV= {Eur. Conf. Comput. Vis.})

@String(NIPS= {Adv. Neural Inform. Process. Syst.})

@String(TOG= {ACM Trans. Graph.})

@String(TIP  = {IEEE Trans. Image Process.})

@String(ICLR = {Int. Conf. Learn. Represent.})

@String(AAAI = {AAAI})

@String(PAMI  = {IEEE TPAMI})

@String(CVPR  = {CVPR})

@String(ICCV  = {ICCV})

@String(ECCV  = {ECCV})

@String(NIPS  = {NeurIPS})

@String(TOG   = {ACM TOG})

@String(TIP   = {IEEE TIP})

@String(ICLR  = {ICLR})

@inproceedings{tang2024lgm,
  title={Lgm: Large multi-view gaussian model for high-resolution 3d content creation},
  author={Tang, Jiaxiang and Chen, Zhaoxi and Chen, Xiaokang and Wang, Tengfei and Zeng, Gang and Liu, Ziwei},
  booktitle=ECCV,
  pages={1--18},
  year={2024}
}

@article{kerbl2023_3dgs,
  title={3d gaussian splatting for real-time radiance field rendering},
  author={Kerbl, Bernhard and Kopanas, Georgios and Leimk{\"u}hler, Thomas and Drettakis, George},
  journal=TOG,
  volume={42},
  number={4},
  pages={1--14},
  year={2023},
  publisher={ACM New York, NY, USA}
}

@inproceedings{ye2025nopose,
  title={No pose, no problem: Surprisingly simple 3d gaussian splats from sparse unposed images},
  author={Ye, Botao and Liu, Sifei and Xu, Haofei and Li, Xueting and Pollefeys, Marc and Yang, Ming-Hsuan and Peng, Songyou},
  booktitle=ICLR,
  pages={54009--54033},
  year={2025}
}

@inproceedings{wewer2024latentsplat,
  title={latentsplat: Autoencoding variational gaussians for fast generalizable 3d reconstruction},
  author={Wewer, Christopher and Raj, Kevin and Ilg, Eddy and Schiele, Bernt and Lenssen, Jan Eric},
  booktitle=ECCV,
  pages={456--473},
  year={2024}
}

@inproceedings{szymanowicz2024splatter,
  title={Splatter image: Ultra-fast single-view 3d reconstruction},
  author={Szymanowicz, Stanislaw and Rupprecht, Chrisitian and Vedaldi, Andrea},
  booktitle=CVPR,
  pages={10208--10217},
  year={2024}
}

@inproceedings{charatan2023pixelsplat,
  title={pixelsplat: 3d gaussian splats from image pairs for scalable generalizable 3d reconstruction},
  author={Charatan, David and Li, Sizhe Lester and Tagliasacchi, Andrea and Sitzmann, Vincent},
  booktitle=CVPR,
  pages={19457--19467},
  year={2024}
}

@inproceedings{liu2024mvsgaussian,
  title={Mvsgaussian: Fast generalizable gaussian splatting reconstruction from multi-view stereo},
  author={Liu, Tianqi and Wang, Guangcong and Hu, Shoukang and Shen, Liao and Ye, Xinyi and Zang, Yuhang and Cao, Zhiguo and Li, Wei and Liu, Ziwei},
  booktitle=ECCV,
  pages={37--53},
  year={2024}
}

@inproceedings{chen2024mvsplat,
  title={Mvsplat: Efficient 3d gaussian splatting from sparse multi-view images},
  author={Chen, Yuedong and Xu, Haofei and Zheng, Chuanxia and Zhuang, Bohan and Pollefeys, Marc and Geiger, Andreas and Cham, Tat-Jen and Cai, Jianfei},
  booktitle=ECCV,
  pages={370--386},
  year={2024}
}

@article{vaswani2017attention,
  title={Attention is all you need},
  author={Vaswani, Ashish and Shazeer, Noam and Parmar, Niki and Uszkoreit, Jakob and Jones, Llion and Gomez, Aidan N and Kaiser, {\L}ukasz and Polosukhin, Illia},
  journal=NIPS,
  volume={30},
  pages={5998--6008},
  year={2017}
}

@inproceedings{zheng2024gpsgaussian,
  title={GPS-Gaussian: Generalizable Pixel-wise 3D Gaussian Splatting for Real-time Human Novel View Synthesis},
  author={Zheng, Shunyuan and Zhou, Boyao and Shao, Ruizhi and Liu, Boning and Zhang, Shengping and Nie, Liqiang and Liu, Yebin},
  booktitle=CVPR,
  pages={19680--19690},
  year={2024}
}

@inproceedings{saito2019pifu,
  title={Pifu: Pixel-aligned implicit function for high-resolution clothed human digitization},
  author={Saito, Shunsuke and Huang, Zeng and Natsume, Ryota and Morishima, Shigeo and Kanazawa, Angjoo and Li, Hao},
  booktitle=ICCV,
  pages={2304--2314},
  year={2019}
}

@inproceedings{schonberger2016colmap,
  title={Structure-from-motion revisited},
  author={Schonberger, Johannes L and Frahm, Jan-Michael},
  booktitle=CVPR,
  pages={4104--4113},
  year={2016}
}

@inproceedings{lipson2021raft-stereo,
  title={Raft-stereo: Multilevel recurrent field transforms for stereo matching},
  author={Lipson, Lahav and Teed, Zachary and Deng, Jia},
  booktitle={3DV},
  pages={218--227},
  year={2021}
}

@inproceedings{peng2021neural-body,
  title={Neural body: Implicit neural representations with structured latent codes for novel view synthesis of dynamic humans},
  author={Peng, Sida and Zhang, Yuanqing and Xu, Yinghao and Wang, Qianqian and Shuai, Qing and Bao, Hujun and Zhou, Xiaowei},
  booktitle=CVPR,
  pages={9054--9063},
  year={2021}
}

@article{loper2015smpl,
  title={SMPL: A skinned multi-person linear model},
  author={Loper, Matthew and Mahmood, Naureen and Romero, Javier and Pons-Moll, Gerard and Black, Michael J},
  journal=TOG,
  volume={34},
  number={6},
  pages={1--16},
  year={2015},
  publisher={ACM New York, NY, USA}
}

@article{wang2004ssim,
  title={Image quality assessment: from error visibility to structural similarity},
  author={Wang, Zhou and Bovik, Alan C and Sheikh, Hamid R and Simoncelli, Eero P},
  journal=TIP,
  volume={13},
  number={4},
  pages={600--612},
  year={2004},
  publisher={IEEE}
}

@inproceedings{zhang2018lpips,
  title={The unreasonable effectiveness of deep features as a perceptual metric},
  author={Zhang, Richard and Isola, Phillip and Efros, Alexei A and Shechtman, Eli and Wang, Oliver},
  booktitle=CVPR,
  pages={586--595},
  year={2018}
}

@misc{twindom,
  Author = {{Twindom}},
  Note = {\url{https://web.twindom.com}},
  Year = {2020}
}

@inproceedings{adamw,
  title={Decoupled Weight Decay Regularization},
  author={Loshchilov, Ilya and Hutter, Frank},
  booktitle=ICLR,
  year={2018}
}

@inproceedings{hu2024gaussianavatar,
  title={Gaussianavatar: Towards realistic human avatar modeling from a single video via animatable 3d gaussians},
  author={Hu, Liangxiao and Zhang, Hongwen and Zhang, Yuxiang and Zhou, Boyao and Liu, Boning and Zhang, Shengping and Nie, Liqiang},
  booktitle=CVPR,
  pages={634--644},
  year={2024}
}

@inproceedings{tu2024tele,
  title={Tele-Aloha: A Telepresence System with Low-budget and High-authenticity Using Sparse RGB Cameras},
  author={Tu, Hanzhang and Shao, Ruizhi and Dong, Xue and Zheng, Shunyuan and Zhang, Hao and Chen, Lili and Wang, Meili and Li, Wenyu and Ma, Siyan and Zhang, Shengping and others},
  booktitle={SIGGRAPH},
  pages={1--12},
  year={2024}
}

@inproceedings{kwon2024generalizable,
  title={Generalizable human gaussians for sparse view synthesis},
  author={Kwon, Youngjoong and Fang, Baole and Lu, Yixing and Dong, Haoye and Zhang, Cheng and Carrasco, Francisco Vicente and Mosella-Montoro, Albert and Xu, Jianjin and Takagi, Shingo and Kim, Daeil and others},
  booktitle=ECCV,
  pages={451--468},
  year={2024}
}

@article{huang2025echoes,
  title={Echoes of the Coliseum: Towards 3D Live streaming of Sports Events},
  author={Huang, Junkai and Subhajyoti Mallick, Saswat and Amat, Alejandro and Ruiz Olle, Marc and Mosella-Montoro, Albert and Kerbl, Bernhard and Vicente Carrasco, Francisco and De la Torre, Fernando},
  journal=TOG,
  volume={44},
  number={4},
  pages={1--17},
  year={2025},
  publisher={ACM New York, NY, USA}
}

@inproceedings{sun2025generalizable,
  title={Generalizable and relightable gaussian splatting for human novel view synthesis},
  author={Sun, Yipengjing and Zhang, Shengping and Wang, Chenyang and Zheng, Shunyuan and Li, Zonglin and Ji, Xiangyang},
  booktitle={SIGGRAPH},
  pages={1--11},
  year={2026}
}

@inproceedings{mescheder2026sharp,
  title={Sharp monocular view synthesis in less than a second},
  author={Mescheder, Lars and Dong, Wei and Li, Shiwei and Bai, Xuyang and Santos, Marcel and Hu, Peiyun and Lecouat, Bruno and Zhen, Mingmin and Delaunoy, Ama{\"e}l and Fang, Tian and others},
  booktitle=ICLR,
  pages={24192--24230},
  year={2026}
}

@inproceedings{xu2025depthsplat,
  title={Depthsplat: Connecting gaussian splatting and depth},
  author={Xu, Haofei and Peng, Songyou and Wang, Fangjinhua and Blum, Hermann and Barath, Daniel and Geiger, Andreas and Pollefeys, Marc},
  booktitle=CVPR,
  pages={16453--16463},
  year={2025}
}

@inproceedings{zhou2026splat,
  title={Splat-SAP: Feed-Forward Gaussian Splatting for Human-Centered Scene with Scale-Aware Point Map Reconstruction},
  author={Zhou, Boyao and Zheng, Shunyuan and Liao, Zhanfeng and Ma, Zihan and Tu, Hanzhang and Liu, Boning and Liu, Yebin},
  booktitle=AAAI,
  pages={13566--13574},
  year={2026}
}

@inproceedings{hong2024lrm,
  title={Lrm: Large reconstruction model for single image to 3d},
  author={Hong, Yicong and Zhang, Kai and Gu, Jiuxiang and Bi, Sai and Zhou, Yang and Liu, Difan and Liu, Feng and Sunkavalli, Kalyan and Bui, Trung and Tan, Hao},
  booktitle=ICLR,
  pages={50678--50702},
  year={2024}
}

@inproceedings{liu2025monosplat,
  title={MonoSplat: Generalizable 3D Gaussian splatting from monocular depth foundation models},
  author={Liu, Yifan and Fan, Keyu and Yu, Weihao and Li, Chenxin and Lu, Hao and Yuan, Yixuan},
  booktitle=CVPR,
  pages={21570--21579},
  year={2025}
}

@inproceedings{wang2024dust3r,
  title={Dust3r: Geometric 3d vision made easy},
  author={Wang, Shuzhe and Leroy, Vincent and Cabon, Yohann and Chidlovskii, Boris and Revaud, Jerome},
  booktitle=CVPR,
  pages={20697--20709},
  year={2024}
}

@inproceedings{leroy2024mast3r,
  title={Grounding image matching in 3d with mast3r},
  author={Leroy, Vincent and Cabon, Yohann and Revaud, J{\'e}r{\^o}me},
  booktitle=ECCV,
  pages={71--91},
  year={2024}
}

@inproceedings{wang2025vggt,
  title={Vggt: Visual geometry grounded transformer},
  author={Wang, Jianyuan and Chen, Minghao and Karaev, Nikita and Vedaldi, Andrea and Rupprecht, Christian and Novotny, David},
  booktitle=CVPR,
  pages={5294--5306},
  year={2025}
}

@inproceedings{wang2026pi,
  title={$\pi^3$: Permutation-Equivariant Visual Geometry Learning},
  author={Wang, Yifan and Zhou, Jianjun and Zhu, Haoyi and Chang, Wenzheng and Zhou, Yang and Li, Zizun and Chen, Junyi and Pang, Jiangmiao and Shen, Chunhua and He, Tong},
  booktitle=ICLR,
  pages={10481--10497},
  year={2026}
}

@inproceedings{lin2026depth,
  title={Depth anything 3: Recovering the visual space from any views},
  author={Lin, Haotong and Chen, Sili and Liew, Junhao and Chen, Donny Y and Li, Zhenyu and Shi, Guang and Feng, Jiashi and Kang, Bingyi},
  booktitle=ICLR,
  pages={141261--141285},
  year={2026}
}

@inproceedings{liu2026worldmirror,
  title={WorldMirror: Universal 3D World Reconstruction with Any-Prior Prompting},
  author={Liu, Yifan and Min, Zhiyuan and Wang, Zhenwei and Wu, Junta and Wang, Tengfei and Yuan, Yixuan and Luo, Yawei and Guo, Chunchao},
  booktitle={ICML},
  year={2026}
}

@inproceedings{zhao2026structsplat,
  title={StructSplat: Generalizable 3D Gaussian Splatting from Uncalibrated Sparse Views},
  author={Zhao, Jia-Chen and Chen, Beiqi and Chen, Xinyang and Wang, Guangcong and Nie, Liqiang},
  booktitle=ECCV,
  year={2026}
}

@article{tyszkiewicz2020disk,
  title={Disk: Learning local features with policy gradient},
  author={Tyszkiewicz, Micha{\l} and Fua, Pascal and Trulls, Eduard},
  journal=NIPS,
  volume={33},
  pages={14254--14265},
  year={2020}
}

@inproceedings{lindenberger2023lightglue,
  title={Lightglue: Local feature matching at light speed},
  author={Lindenberger, Philipp and Sarlin, Paul-Edouard and Pollefeys, Marc},
  booktitle=ICCV,
  pages={17627--17638},
  year={2023}
}

@article{umeyama1991least,
  title={Least-squares estimation of transformation parameters between two point patterns},
  author={Umeyama, Shinji},
  journal=PAMI,
  volume={13},
  number={4},
  pages={376--380},
  year={1991},
  publisher={IEEE}
}

@inproceedings{hu2025eva,
  title={EVA-Gaussian: 3D Gaussian-based Real-time Human Novel View Synthesis under Diverse Multi-view Camera Settings},
  author={Hu, Yingdong and Liu, Zhening and Shao, Jiawei and Lin, Zehong and Zhang, Jun},
  booktitle={ICCVW},
  pages={2634-2643},
  year={2025}
}

@inproceedings{shi2016pixshuff,
  title={Real-time single image and video super-resolution using an efficient sub-pixel convolutional neural network},
  author={Shi, Wenzhe and Caballero, Jose and Husz{\'a}r, Ferenc and Totz, Johannes and Aitken, Andrew P and Bishop, Rob and Rueckert, Daniel and Wang, Zehan},
  booktitle=CVPR,
  pages={1874--1883},
  year={2016}
}

@inproceedings{sun2023pixunshuff,
  title={Hybrid pixel-unshuffled network for lightweight image super-resolution},
  author={Sun, Bin and Zhang, Yulun and Jiang, Songyao and Fu, Yun},
  booktitle=AAAI,
  pages={2375--2383},
  year={2023}
}

@article{zhou2025gps,
  title={Gps-gaussian+: Generalizable pixel-wise 3d gaussian splatting for real-time human-scene rendering from sparse views},
  author={Zhou, Boyao and Zheng, Shunyuan and Tu, Hanzhang and Shao, Ruizhi and Liu, Boning and Zhang, Shengping and Nie, Liqiang and Liu, Yebin},
  journal=PAMI,
  year={2025},
  publisher={IEEE}
}

@inproceedings{yang2024depthanything,
  title={Depth Anything: Unleashing the Power of Large-Scale Unlabeled Data},
  author={Yang, Lihe and Kang, Bingyi and Huang, Zilong and Xu, Xiaogang and Feng, Jiashi and Zhao, Hengshuang},
  booktitle=CVPR,
  pages={10371--10381},
  year={2024}
}

@inproceedings{yang2025fast3r,
  title={Fast3R: Towards 3D Reconstruction of 1000+ Images in One Forward Pass},
  author={Yang, Jianing and Sax, Alexander and Liang, Kevin J and Henaff, Mikael and Tang, Hao and Cao, Ang and Chai, Joyce and Meier, Franziska and Feiszli, Matt},
  booktitle=CVPR,
  pages={21924--21935},
  year={2025}
}

@inproceedings{wang2025cut3r,
  title={Continuous 3D Perception Model with Persistent State},
  author={Wang, Qianqian and Zhang, Yifei and Holynski, Aleksander and Efros, Alexei A and Kanazawa, Angjoo},
  booktitle=CVPR,
  pages={10510--10522},
  year={2025}
}

@inproceedings{zhang2025monst3r,
  title={Monst3r: A simple approach for estimating geometry in the presence of motion},
  author={Zhang, Junyi and Herrmann, Charles and Hur, Junhwa and Jampani, Varun and Cole, Forrester and Sun, Deqing and Yang, Ming-Hsuan and others},
  booktitle=ICLR,
  pages={82863--82886},
  year={2025}
}

@article{smart2024splatt3r,
  title={Splatt3R: Zero-shot Gaussian Splatting from Uncalibrated Image Pairs},
  author={Smart, Brandon and Zheng, Chuanxia and Laina, Iro and Prisacariu, Victor Adrian},
  journal={arXiv},
  year={2024}
}

@article{jiang2025anysplat,
  title={Anysplat: Feed-forward 3d gaussian splatting from unconstrained views},
  author={Jiang, Lihan and Mao, Yucheng and Xu, Linning and Lu, Tao and Ren, Kerui and Jin, Yichen and Xu, Xudong and Yu, Mulin and Pang, Jiangmiao and Zhao, Feng and others},
  journal=TOG,
  volume={44},
  number={6},
  pages={1--16},
  year={2025},
  publisher={ACM New York, NY, USA}
}

@inproceedings{tu2025gbc,
  title={GBC-Splat: Generalizable Gaussian-Based Clothed Human Digitalization under Sparse RGB Cameras},
  author={Tu, Hanzhang and Liao, Zhanfeng and Zhou, Boyao and Zheng, Shunyuan and Zhou, Xilong and Zhang, Liuxin and Wang, QianYing and Liu, Yebin},
  booktitle=CVPR,
  pages={26377--26387},
  year={2025}
}

@inproceedings{jiang2026hireff,
  title={HiReFF: High-Resolution Feedforward Human Reconstruction from Uncalibrated Sparse-View Video},
  author={Jiang, Yiming and Tu, Hanzhang and Song, Wenfeng and Lin, Siyou and An, Liang and Li, Shuai and Hao, Aimin and Liu, Yebin},
  booktitle=ECCV,
  year={2026}
}

@article{zhang2025ins,
  title={Ins-hoi: Instance aware human-object interactions recovery},
  author={Zhang, Jiajun and Zhang, Yuxiang and Zhang, Hongwen and Zhou, Xiao and Zhou, Boyao and Shao, Ruizhi and Hu, Zonghai and Liu, Yebin},
  journal=PAMI,
  volume={47},
  number={11},
  pages={9655--9672},
  year={2025},
  publisher={IEEE}
}

@inproceedings{yu2021function4d,
  title={Function4d: Real-time human volumetric capture from very sparse consumer rgbd sensors},
  author={Yu, Tao and Zheng, Zerong and Guo, Kaiwen and Liu, Pengpeng and Dai, Qionghai and Liu, Yebin},
  booktitle=CVPR,
  pages={5742--5752},
  year={2021}
}

\end{document}